\documentclass[10pt,twocolumn,letterpaper]{article}

\usepackage{iccv}
\usepackage{times}
\usepackage{epsfig}
\usepackage{graphicx}
\usepackage{amsmath}
\usepackage{amssymb}
\usepackage{multirow}
\usepackage{cuted}
\usepackage{capt-of}
\usepackage{tabularx}
\usepackage{booktabs}
\usepackage{authblk}
\usepackage{xcolor}
\usepackage[accsupp]{axessibility}  

\usepackage[breaklinks=true,bookmarks=false]{hyperref}

\iccvfinalcopy 

\def\iccvPaperID{6297} 

\ificcvfinal\fi

\definecolor{ecyan}{rgb}{0.1,0.6,0.5}

\begin{document}

\title{Collaging Class-specific GANs for Semantic Image Synthesis}

\author{
\vspace{-0.25in}
Yuheng Li\textsuperscript{1} \quad 
Yijun Li\textsuperscript{2} \quad 
Jingwan Lu\textsuperscript{2} \quad 
Eli Shechtman\textsuperscript{2} \quad 
Yong Jae Lee\textsuperscript{1} \quad 
Krishna Kumar Singh\textsuperscript{2}  
\\
\vspace{-0.1in}
\textsuperscript{1}University of Wisconsin-Madison \quad \textsuperscript{2}Adobe Research
\\

}

\ificcvfinal\thispagestyle{empty}\fi

\twocolumn[{%
	\maketitle
	\vspace{-1.25cm}
	\renewcommand\twocolumn[1][]{#1}%
	\begin{center}
		\centering
		\includegraphics[width=0.95\textwidth]{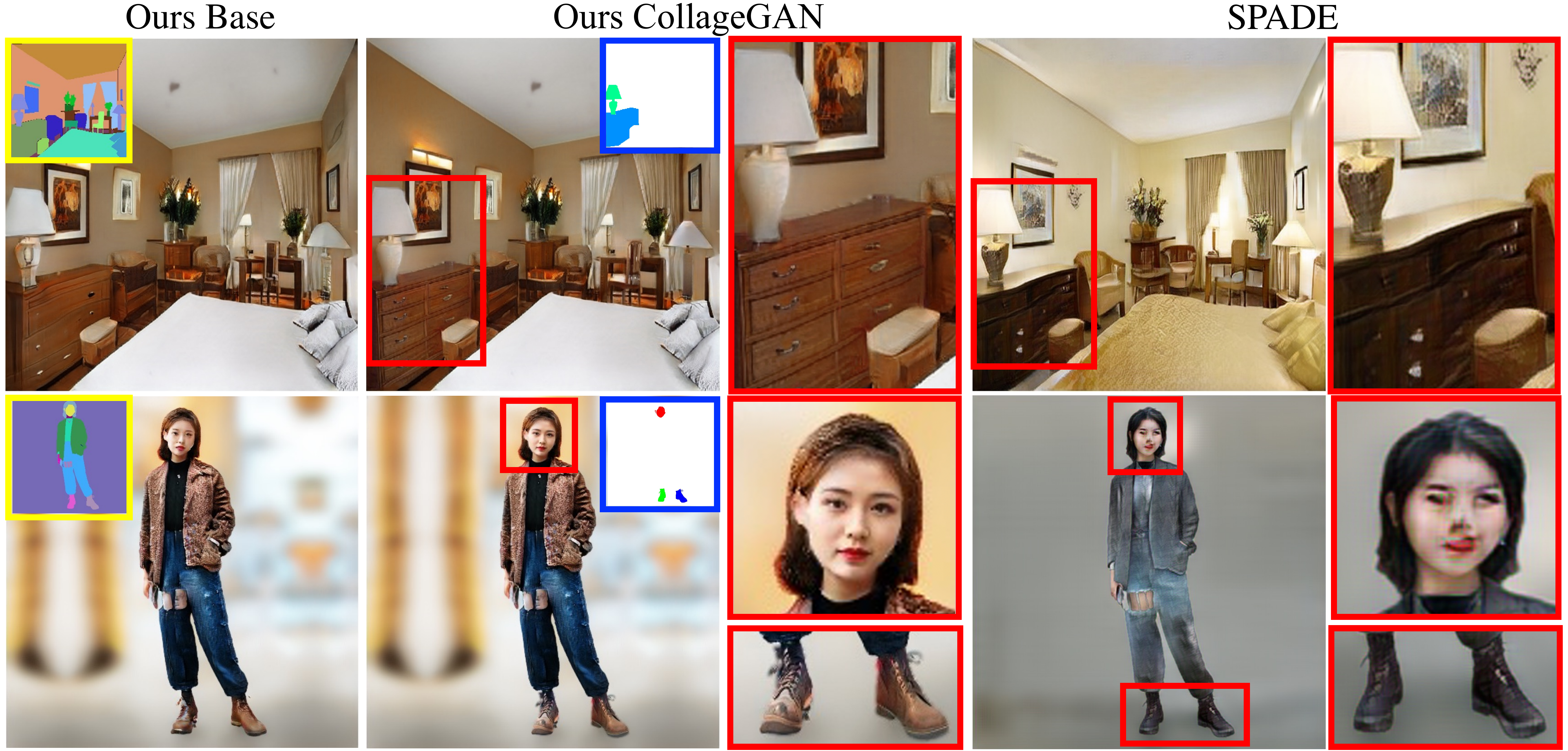}
		\captionof{figure}{High resolution image generation conditioned on a segmentation map (yellow inset). Our base model (1st column) generates more realistic results than SPADE~\cite{park-cvpr2019} (4-5th column). We further improve the quality of results by using our class-specific generators to generate foreground objects or parts and compose them on the image generated by our base model. The segmentation maps in 2nd column (blue insets) show the parts that are modified by our bank of GANs and the zoomed-in results are shown next to each image. }
		\label{fig:teaser}
	\end{center}%
}]

\begin{abstract}
\vspace{-0.15in}
We propose a new approach for high resolution semantic image synthesis. It consists of one base image generator and multiple class-specific generators. The base generator generates high quality images based on a segmentation map. To further improve the quality of different objects, we create a bank of Generative Adversarial Networks (GANs) by separately training class-specific models. This has several benefits including -- dedicated weights for each class; centrally aligned data for each model; additional training data from other sources, potential of higher resolution and quality; and easy manipulation of a specific object in the scene. Experiments show that our approach can generate high quality images in high resolution while having flexibility of object-level control by using class-specific generators.
Project page: \url{https://yuheng-li.github.io/CollageGAN/}
\end{abstract}
\vspace{-0.22in}

\section{Introduction}

Image generation has been explored extensively in both the unconditional~\cite{goodfellow-gantutorial2016,karras-iclr2018,karras-cvpr2019,karras-cvpr2020} and conditional~\cite{isola-pix2pix2017,zhang-iccv17,wang-cvpr2018,Johnson-cvpr2018,brock-iclr2019,park-cvpr2019} settings.
In the unconditional setting, an image is generated by randomly sampling a latent code. With the advent of StyleGAN2~\cite{karras-cvpr2020}, for some classes such as faces, the generated images are almost indistinguishable from real images. 
In the conditional setting, an image is generated based on some input conditioning signal such as another image, a segmentation map, or other priors. Most notably, SPADE~\cite{park-cvpr2019} (i.e., spatially-adaptive normalization)  significantly boosted image quality for semantic image generation. This normalization and its variants have been used as standard building blocks in many following works~\cite{liu-nips2019,Zhu-cvpr2020,tang-cvpr2020,Sushko-iclr2021}. However, the image quality of conditional GANs is still inferior to that of unconditional GANs, especially StyleGAN2. Also, most approaches output $256 \times 256$ resolution images, which is not high enough for many real world applications, and their generation quality is inferior for higher resolutions.

In this work, we target high resolution image generation conditioned on input segmentation maps (yellow insets in Fig.~\ref{fig:teaser}, first column). 
Contrary to most prior work which typically follow the SPADE architecture design, we explore a new direction. First, we develop a conditional version of StyleGAN2.  Our generator backbone builds upon the original StyleGAN2 architecture, which can generate better quality images and is more easily scaled to high resolutions. Specifically, we use an encoder to extract hierarchical feature representations from the input segmentation map and use it to modulate the StyleGAN2 generator. The resulting generator is less memory intensive and faster to train compared to SPADE based approaches.

Second, instead of using a single generator, we use \emph{multiple class-specific generators} to improve the quality of small foreground objects. There are several advantages of using different models for different classes: 
(i) Class-specific generators with dedicated weights learned for each foreground class have the capability to generate more details. Imagine a bedroom filled with objects of different scales such as bed, lamp, table, chair, chest and others (Fig.~\ref{fig:teaser}, first row). A single GAN might allocate most of its capacity toward generating larger content such as floor, walls and bed since they contribute most to the overall realism, and therefore neglect the details of smaller objects such as lamps and chests. Using separate generators for smaller objects result in more textural details and better shape integrity in those regions (Fig.~\ref{fig:teaser}, columns 2-3).
(ii) Class-specific generators can be trained with image crops that always align the objects to the center. As shown in~\cite{karras-cvpr2019, karras-cvpr2020}, spatial alignment is often crucial for high quality generation. On the other hand, a single GAN generating an entire scene at once needs to deal with objects appearing in arbitrary locations which is more difficult to learn. 
(iii) Class-specific generators can benefit from more training data, which is especially important for more rare classes. Other than the scene training set from which all foreground objects are extracted, one can leverage separate class-specific datasets for training each foreground generator. 
(iv) Class-specific generators enable more applications. For example, we can generate out-of-distribution scene images such as a car on a sidewalk (Fig.~\ref{fig:ood}), and we can also easily change the appearance of one object instance in the scene without modifying the rest of it (Fig.~\ref{fig:replace_real}).

However, na\"ively training separate generators and combining their results does not generate satisfactory results. It can lead to appearance inconsistency among generated objects in the same scene; e.g., a generated foreground object might not have compatible colors and perspective with its background or in the context of other objects. Training all models together based on shared global information will address the inconsistency issue, but it is extremely computationally demanding and cannot scale. Instead, we provide the result of our single GAN base model as input to each class-specific generator since it captures the global knowledge of the scene and constrains the foreground object appearance to be harmonious with each other. Although our focus is on generating complex scenes, our approach can also be used to generate a single object class with multiple complex parts like humans (Fig.~\ref{fig:teaser}, second row). We demonstrate this by training separate class-specific generators for faces and shoes.

\vspace{0.5em}
\noindent\textbf{Contributions.} (1) A general framework for high quality conditional generation of complex scenes by training multiple class-specific generators; (2) A powerful StyleGAN2-based conditional base generator; (3) We demonstrate state-of-the-art image quality synthesized by our model against existing methods and show some potential applications.

\vspace{-0.5em}

\section{Related work}

\noindent\textbf{High Resolution Semantic Image Generation.} GANs \cite{goodfellow-nips2014} have  become one of the most promising models to generate photorealistic images. Many different architectures~\cite{chen-nips16, zhu-iccv2017,zhang-iccv17,wang-cvpr2018,karras-iclr2018,brock-iclr2019,karras-cvpr2019,karras-cvpr2020} and training techniques~\cite{arjovsky-icml2017,gulrajani-nips17,Miyato-iclr2018} have been proposed to explore different tasks~\cite{chen-nips16, zhu-iccv2017,zhang-iccv17,wang-cvpr2018}, improve generation quality~\cite{brock-iclr2019,karras-cvpr2019,karras-cvpr2020}, and stabilize training~\cite{arjovsky-icml2017,gulrajani-nips17,Miyato-iclr2018}. There are also many great works exploring conditional image generation~\cite{zhang-iccv17,wang-cvpr2018,Johnson-cvpr2018,brock-iclr2019,park-cvpr2019,tang-cvpr2020,Sushko-iclr2021}. In particular, segmentation map to image generation~\cite{wang-cvpr2018,park-cvpr2019,tang-cvpr2020,Sushko-iclr2021} is popular due to its wide applications like flexible content creation for image editing. Although these works generate impressive results at lower resolutions, their higher resolution results for complex settings such as indoor scenes or full human bodies are often inferior and less detailed. Even at lower resolutions, the quality of foreground objects can be low due to the model focusing more on the background, which typically has more pixels than the foreground. The current
state-of-the-art method for high resolution image generation is StyleGAN2~\cite{karras-cvpr2019,karras-cvpr2020}. However, it is an unconditional model which lacks controllability over the image generation process. Some recent concurrent works~\cite{lewis-arxiv2020,sarkar-arxiv2021} modify the StyleGAN2 architecture for conditional image generation for the specific domain of human bodies. In contrast, we propose a general purpose approach to condition StyleGAN2 on segmentation map inputs of various scene categories (e.g., bedrooms and street scenes) and complex objects (e.g., humans).       

\vspace{0.5em}
\noindent\textbf{Image generation through unshared weights.} Most image generation models have a monolithic architecture~\cite{wang-cvpr2018,karras-iclr2018,brock-iclr2019,park-cvpr2019,karras-cvpr2019,karras-cvpr2020}, which outputs a final image without specialized weights for different classes or semantic regions. Others~\cite{yang-iclr17,huang-iccv2017,li-cvpr2018,gu-cvpr2019,hinz-iclr019,singh-cvpr2019,tang-cvpr2020} model an image via different components, e.g., by having global and local branches~\cite{huang-iccv2017,li-cvpr2018,gu-cvpr2019, hinz-iclr019} with the local branch being specialized for different objects/parts. However, these models are usually face specific~\cite{huang-iccv2017,li-cvpr2018,gu-cvpr2019}. The work of \cite{hinz-iclr019, yang-iclr17,singh-cvpr2019} propose to separate background and foreground generation. But they can only generate single object images and do not have dedicated generators for specific foreground categories. \cite{tang-cvpr2020} shares a similar idea with us. However, instead of having separate modules for different classes, it only has a few separate convolution layers for each class at the end of the generator, which means that it is deprived of our advantages of aligning training data, having an entire network dedicated to a class, and using data from other sources. Moreover, the model is very memory extensive and time consuming to train, as all the classes have to be trained together in an end-to-end fashion. In contrast, our class-specific generators can be trained for selected classes independently on separate GPUs in parallel.     

\section{Approach}

We aim to generate high-quality scene images based on input segmentation maps by using a powerful base generator that generates an entire scene without differentiating objects and parts, and multiple class-specific generators that improve the appearances of small foreground objects. Both types of generators use the same architecture design with minor changes. We first introduce our new architecture (Section~\ref{sec:arc}). Then, we describe our training (Section~\ref{sec:training}) and inference pipelines (Section~\ref{sec:inference}).

\subsection{Architecture}
\label{sec:arc}

\begin{figure}[t!]
    \centering
    \includegraphics[width=0.48\textwidth]{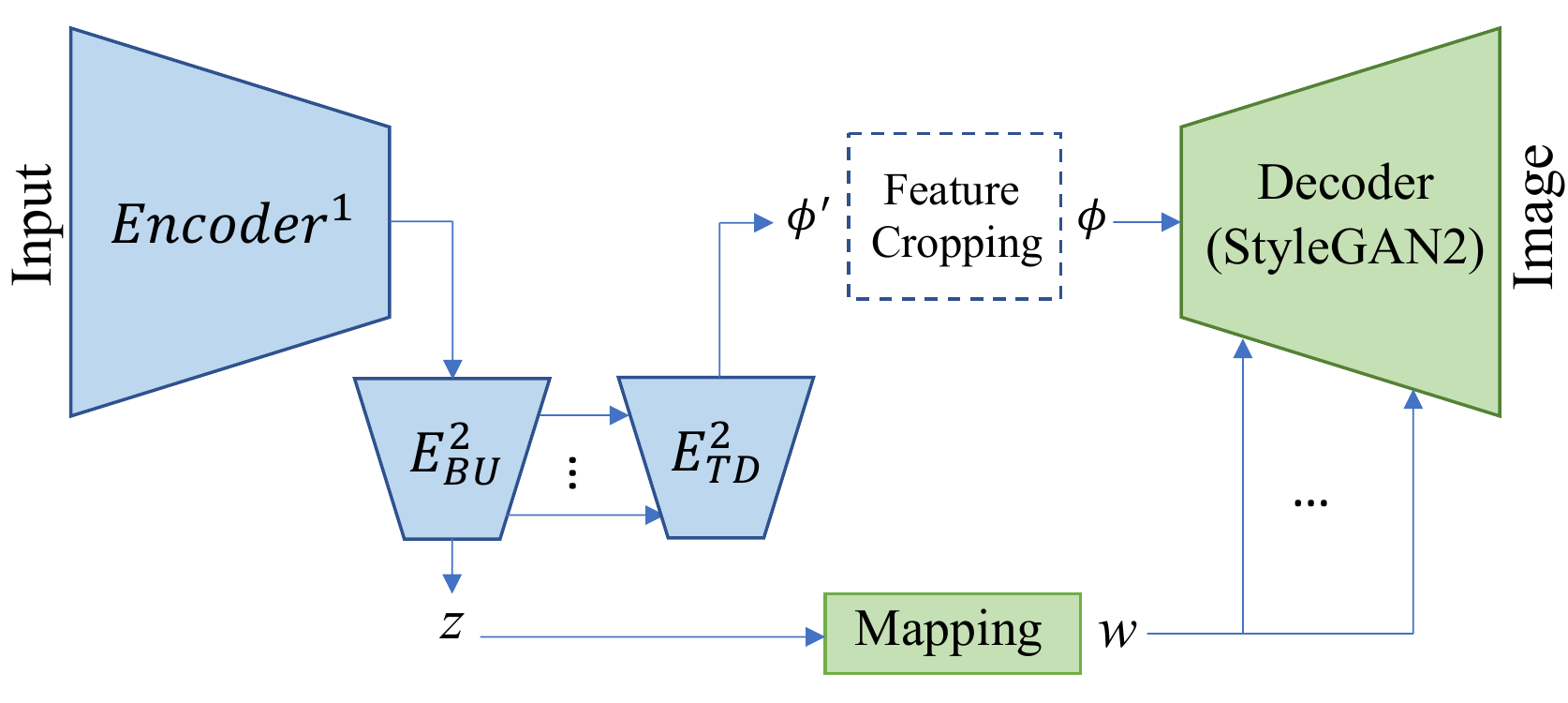}
    \vspace{-0.01in}
    \caption{\textbf{The common architecture used for both the base model and the class-specific models} We modify the StyleGAN2 architecture by replacing its input constant with the output from the encoder. $z$ is also conditioned on the encoder output. }
    \label{fig:architecture}
\end{figure}

Fig.~\ref{fig:architecture} shows our architecture, which is used by both our class-specific models and the base model. It consists of an encoder (blue module) and a decoder (green module). The encoder takes in input (e.g., a semantic and instance edge map in our base model, more details of the content of the input will be introduced in the section~\ref{sec:training}) and processes it into a latent code $z$ and a spatial feature tensor $\phi\prime$. The $\phi\prime$ is next processed by the feature cropping model and fed into decoder together with $z$ to synthesize realistic images. Next, we will introduce details of the encoder, the feature cropping, and the decoder.

In the encoder, we first have a several of conv and downsampling layers ($Encoder^1$) , then we extend its backbone at the last $N$ layers (refer $E^2_{BU}$ and $E^2_{TD}$ in the Fig.~\ref{fig:architecture}) with a feature pyramid~\cite{lin-cvpr2017}, enhancing higher resolution features (that are more accurately localized to the input) with lower resolution features (that are more semantic and has larger receptive field to capture global information). For example, if the input was a segmentation map then the higher resolution features would be more aligned with the input layout whereas the lower resolution features would have more global information about all the classes present in the segmentation map. As shown in the Fig.~\ref{fig:architecture}, the second part of the encoder, $E^2$, consists of two pathways: bottom-up ($E^2_{BU}$) for downsampling and top-down ($E^2_{TD}$) for upsampling and merging features from $E^2_{BU}$. The output of $E^2_{TD}$ is $\phi\prime$. Note that $z$ is the output of $E^2_{BU}$ without being processed by $E^2_{TD}$.

\begin{figure*}[t!]
    \centering
    \includegraphics[width=0.92\textwidth]{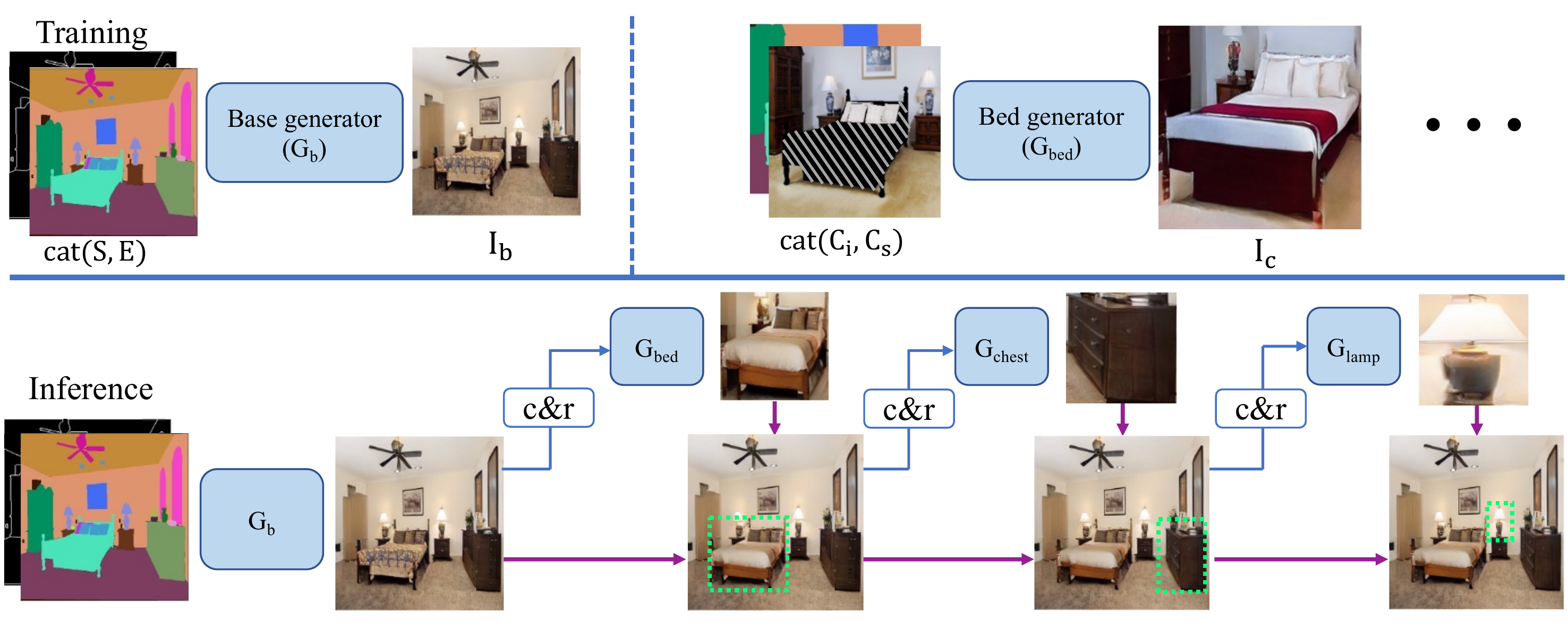}
    \caption{\textbf{Our Approach.} During training, our base generator is trained to generate an entire image, whereas our class-specific generators take in the cropped real image with object information removed (shaded region is filled with either zeros or low frequency information of the object) and cropped segmentation map as context information to generate the instance for the corresponding class. During inference, the base generator first generates the entire image and then the class-specific generators sequentially generate specific regions of the image while considering the previous generations as context. c\&r means the cropping and instance information removing operation. Note that we do not show cropped semantic map as input of class-specific models in the inference pipeline for simplicity.}
    \label{fig:pipeline}
    \vspace{-0.1in}
\end{figure*}

Next, the spatial feature tensor $\phi\prime$ is fed into the feature cropping module, a fixed operation without learnable parameters. In class-specific models, the feature cropping is predefined to crop class instance regions from the $\phi\prime$ to get the starting feature tensor $\phi$ for the decoder. Refer the section~\ref{sec:training} for more explanations. For the base generator, the feature cropping module is identity mapping; i.e., $\phi = \phi\prime$.

The green module in our architecture is the decoder adopted from StyleGAN2~\cite{karras-cvpr2019,karras-cvpr2020}. In StyleGAN2, the generator backbone takes a learnt constant as input. Instead, our generator takes the output from the feature cropping module as input since it provides the generator features that are relevant for the class at hand. As the resolution of $\phi$ is usually higher than the constant used in StyleGAN2~\cite{karras-cvpr2019,karras-cvpr2020}, thus we skip its first K layers. More details of our architecture can be found in the supp.  

\subsection{Training pipeline}
\label{sec:training}

\noindent\textbf{Base generator.}
As shown in the top left of Fig.~\ref{fig:pipeline}, the base generator takes in segmentation map $S$ and instance edge map $E$, to generate a base image $I_b$ that covers the entire scene, i.e., $I_b = G_b( cat(S,E) )$, where $cat(\cdot,\cdot)$ is a channel-wise concatenation. $G_b$ is our base generator including both the encoder and the decoder architectures in Fig.~\ref{fig:architecture}. Using a spatial feature tensor as input to StyleGAN2 rather than a vector~\cite{karras-cvpr2020} gives the model strong guidance on how the generated spatial structure should look like. By sampling different $z$, our model can generate different results given the same segmentation map.

\vspace{0.5em}
\noindent\textbf{Class-specific generator.} To further improve the quality of smaller object classes, we train multiple class-specific generators as shown in the top right of Fig.~\ref{fig:pipeline}. If we generate each object instance without taking into account their context, the generated instances might look great on their own, but the final result after compositing multiple instances may look inharmonious due to inconsistent orientation, color, or lighting among different objects. For example, without context, our lamp-specific generator will generate lamps that do not look compatible with the lighting environment of the scene and shadows on the wall as shown in Fig.~\ref{fig:study_context}. 

To address that, we provide some context information around the target instance as input to our class-specific generators. Specifically, we use the enlarged (2 times in both dimensions) box of an instance to crop both the real image $I_{real\_scene}$ and its segmentation map $S$ to get the cropped real image $C_i$ and the cropped segmentation map $C_s$ for this instance. Next, we use two different ways to remove the instance information from the $C_i$ before providing it for the generation (the shaded region in Fig.~\ref{fig:pipeline} top-right). The first one is to mask out the foreground region with zeros. The second approach is to blur the foreground region to retain only the low frequency information, in which case we hope the generated result will roughly follow the original instance color theme. The $C_{i}$ (with instance information removed) and  $C_s$ are concatenated and used as context $C= cat(C_i,C_s)$ for the class-specific generator $G_c$ to generate a specific instance $I_{c} = G_{c}( C )$. Note that $C_i$ is not cropped from $I_b$, but from the real image during training. The reason is that the real image provides the perfect groundtruth to supervise the hallucination of the foreground object from the context while the generated $I_b$ may contain artifacts. The feature cropping module (See Fig.~\ref{fig:architecture}) inside $G_{c}$ is used to crop the spatial feature $\phi\prime$ to obtain the $\phi$ which is corresponding to the instance region only. The cropping box is calculate such that the final object is tightly generated within $I_c$. In this way, we can fully take advantage of decoder's capacity without generating any context outside the instance box.

To force the model to use $C$, we apply the perceptual loss~\cite{johnson-eccv2016} between the generated instance and the target instance, $I_{real\_ins}$, which is directly cropped from the real image $I_{real\_scene}$ using the instance box without being enlarged. As the background pixels in $I_{real\_ins}$ already exist in $C$ (i.e., $C_i$), the generator will autoencode the background region. In this way, the generator learns to gather hints from the surrounding context of the target instance and generates foreground pixels that look consistent with the background. Also, this loss can help our model maintains the low level frequency information provided in the $C_i$ (the second case of removing instance information). 

\noindent\textbf{Training losses.}
We use the adversarial loss as well  regularizations used in StyleGAN2~\cite{karras-cvpr2020} and refer to all of them combined as $\mathcal{L}_{stylegan}$. For adversarial loss, the real distributions are $ \{I_{real\_scene}\} $ and $\{I_{real\_ins}\}$ for our base and class-specific generator respectively. To regularize our encoder, we apply KL-Divergence~\cite{Kingma-iclr2014} to the output of encoder $z$ forcing it to follow normal distribution to support multi-modal synthesis during inference, $\mathcal{L}_{kl}$. We also use perceptual loss~\cite{johnson-eccv2016}:
 $   \mathcal{L}_{perceptual} = \sum\limits_l || V_l(I_{gen}) - V_l(I_{real}) ||_1$
where $V_l(\cdot)$ is output of $l_{th}$ layer of a pretrained VGG network~\cite{simonyan-iclr2015}. $I_{gen}$ is $I_b$ and $I_{c}$; $I_{real}$ is $I_{real\_scene}$ and $I_{real\_ins}$ in our base and class-specific model, respectively. To summarize our overall training loss is:
 $   \mathcal{L} = \mathcal{L}_{stylegan} + \lambda_1* \mathcal{L}_{kl} + \lambda_2*\mathcal{L}_{perceptual}$
The loss weights and the frequency of regularization within $\mathcal{L}_{stylegan}$ are the same as in StyleGAN2~\cite{karras-cvpr2020}. More details can be found in the supp.

\subsection{Inference pipeline}
\label{sec:inference}
The bottom part of Fig.~\ref{fig:pipeline} describes our entire pipeline during inference. Starting from the input segmentation map and instance edge map, we use our base model to generate the base image $I_b$. Then, the generated base image plays the role of providing context for each class-specific generator. Once an instance is generated by a class-specific generator, its result is composited into the base image. The new composited image becomes the new base image for the next instance. This process continues until all instances are generated. In practice, we generate instances based on their size (largest first) as larger objects are more likely to serve as context for smaller ones.  

For better composition quality, we dilate and soften boundaries of both the generated instances and the instance masks before applying alpha blending. More details about the composition can be found in the supp. 
Though different from training where real images are used to provide context, we find that using the generated base image $I_b$ to provide context at inference time also leads to good results.

\begin{table}[t!]
    \begin{center}
        \scriptsize
        \begin{tabular}{ c|c|c } 
            Scene & Classes & Training data sources \\
            \hline
            \multirow{2}{4em}{Bedroom} & bed, chair, table & (1)+(4) \\ 
            & chest, lamp, pillow & (1)+(4)+(5) \\ 
            \hline
            \multirow{1}{4em}{Human} & shoes, face, upper clothes & (2)  \\ 
            \hline
            \multirow{2}{4em}{Cityscapes} & car & (3)+(6) \\ 
            & person &  (3)+(6)+(7) \\ 
        \end{tabular}
	    \caption{ We can flexibly combine datasets to train our class-specific generators.}
	    \label{table:class_training_data}
	\end{center}
	\vspace{-0.3in}
\end{table}

\section{Experiments}
We perform quantitative and qualitative evaluations comparing our base model and CollageGAN model (combining results of class-specific generators) with prior arts.

\vspace{0.5em}
\noindent\textbf{Datasets.}
For baseline comparisons, we conduct experiments on the following datasets.

\noindent(1) \textbf{Bedroom}. We combine two existing datasets to get 74318 training images in total. We use the training set of $bedroom$ category from ADE20k~\cite{zhou-cvpr2017}. 

Additionally, we use the $bedroom$ and $hotel\_room$ categories from places~\cite{zhou-TPAMI2017} and apply a segmentor~\cite{wu-github2019} trained on ADE20K to get pseudo annotation. We use test set from the $bedroom$ category from ADE20K for evaluation.

\noindent(2) \textbf{Full human body dataset}. We collected 67560 high resolution images on full human body and annotated them with 24 common classes such as faces, upper-clothes, left shoes, right shoes. We randomly select 1/10 of the images as testing images (6757). We blur out the background region to focus the network capacity on generating the human instead of the complex background.

\noindent(3) \textbf{Cityscapes}~\cite{cordts-cvpr2016}. It contains street scene images in German cities. The number of training and testing images are 3000 and 500, respectively.

We use all three datasets to train our base model and baselines.
Our base model usually performs well for the classes of large extent in the scene, for example beds in the bedroom scene or large background categories like walls and floors. Hence, in order to show the impact of our CollageGAN, we train class-specific generators on classes of objects that are usually small and not synthesized well by our base model and baselines. 

Since we have separate models for different classes, we can use the following datasets as extra training data sources for generating bedrooms and cityscapes. For bedroom, we use (4) \textbf{iMaterialist}~\cite{imaterialist} and (5) \textbf{Indoor} dataset ($childs\_room$,  $dining\_room$ and $living\_room$ from places dataset~\cite{zhou-TPAMI2017}). For cityscapes, we use 
(6) \textbf{Cityscapes\_extra}~\cite{cordts-cvpr2016} and (7) \textbf{Caltech Pedestrian}~\cite{Dollar-cvpr2009} dataset. More details about these additional data can be found in the supp. Table~\ref{table:class_training_data} summarizes the classes we choose and their training sources.
We trained all the base models to generate $512 \times 512$ resolution images for bedroom and human dataset, and $1024 \times 512$ images for the cityscapes dataset. Since the resolution of each class varies, the class-specific generators are trained at $128 \times 128$ or $256 \times 256$ depending on average size of each class. For quantitative evaluation, We train all classes, except for person category in cityscapes, with blurred foreground region such that the model can try to maintain the color tone of instances in the base image during inference time. Our base model generates lower quality persons in cityscapes, hence blurred region wasn't very useful and we simply mask instead. More implementation details are in the supp.

\begin{table}[t!]
    \begin{center}
    \scriptsize
        \begin{tabular}{ c|c|c|c|c } 
       
            Datasets & SPADE & OASIS & LGGAN & Ours\\
            \hline
            Bedroom & 45.84 & 39.13 & NA & \textbf{34.41}\\
            
            \hline
            Human & 38.53 & 8.65 & NA & \textbf{7.51}\\
            \hline
            Cityscapes & 59.68  & 50.90 & 61.46  & \textbf{47.07}\\
            
        \end{tabular}

	    \caption{ \textbf{FID (lower is better).} Our base model has consistently lower FID than the baselines. }
	    \label{table:fid_base}

	\end{center}
	\vspace{-0.3in}
\end{table}

\begin{figure*}[t!]
    \centering
    \includegraphics[width=0.95\textwidth]{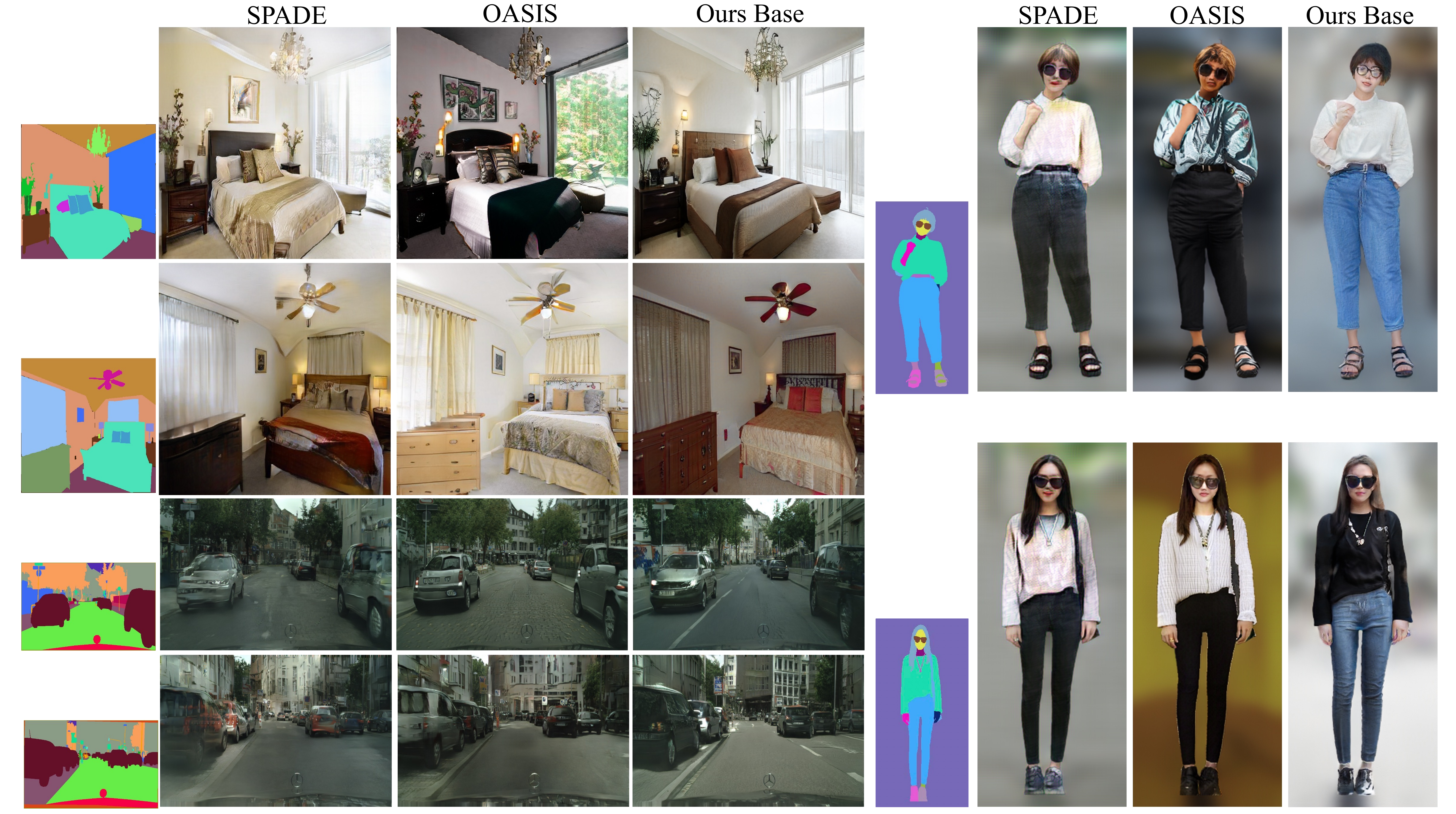}
    \caption{ Visual comparison of segmentation map to image synthesis results on the ADE20K bedroom (512x512), full human body (512x512), and Cityscapes (1024x512) datasets. Our base model generates much more realistic images compared to SPADE and OASIS.}
    \label{fig:baselne_comparison}
    \vspace{-0.1in}
\end{figure*}

\subsection{Base model results}

We compare our base model with SPADE~\cite{park-cvpr2019} and its recent variants LGGAN~\cite{tang-cvpr2020} and OASIS~\cite{Sushko-iclr2021} which are state-of-the-art for the segmentation map to image generation task.  As we are targeting higher resolution than previous approaches ($256 \times 256$ for ADE20K and $512 \times 256$ for cityscapes), we train their models at higher resolution and also provide them instance map for fair comparison. We trained them using their default parameters and verified the result quality of SPADE with the original authors. During training, all SPADE-based approaches are very memory demanding. For example, SPADE and OASIS takes around 16 GB per image to train $512 \times 512$ bedroom images while our model takes only 4 GB. LGGAN cannot fit a single image on a 32 GB V100 GPU for the bedroom dataset as it has a huge number of parameters due to the separate conv layers for each class. We are able to run LGGAN for the datasets with fewer classes like human and cityscapes (takes around 32 GB per image) but training is extremely slow; it would take more than a month for the human dataset with eight 32 GB V100 GPUs. Hence we only compare against LGGAN on cityscapes which is faster to train due to fewer images.

\begin{table}[t!]
    \begin{center}
        \scriptsize
        \begin{tabular}{ c|c|c|c } 
            Datasets & Ours vs SPADE & Ours vs OASIS & Ours vs LGGAN \\
            \hline
            Bedroom & 90.0\% & 73.2\% & NA \\
            \hline
            Human & 82.4\% & 63.2\% & NA \\
            \hline
            Cityscapes & 59.2\% & 35.2\% (83.6\%)  & 62.0\% \\
        \end{tabular}
	    \caption{ Percent of times users preferred our base model over baselines. Our base model is consistently preferred over baselines except for cityscapes when compared with OASIS. But our CollageGAN model is preferred 83.6\% times over OASIS. }
	    \label{table:realism_base}
	\end{center}
\vspace{-0.3in}
\end{table}

\noindent\textbf{Quantitative results and human evaluation.} In Table~\ref{table:fid_base}, we compare FID~\cite{fid} of our base model with all the other approaches. Our base model gets the lowest FID on all three datasets.We also conduct user studies on Amazon Mechanical Turk
(AMT). Specifically, we show the segmentation map and two generated images (ours vs one baseline) side-by-side to the AMT workers. We employ a two-alternative forced choice option for workers by forcing them to choose one image which looks more realistic. We showed 250 image pairs for each comparison and asked 5 unique workers to judge each pair. Table~\ref{table:realism_base} shows our human evaluation results. Users strongly favored our approach over the baselines except for OASIS on the cityscapes dataset. The reason is that our base model sometimes generates less details for smaller objects when the training set is small (cityscapes has only has 3000 images). We can overcome this issue by using our CollageGAN consisting of class-specific generators for the smaller objects like car and person in the cityscape dataset. We then compared the results of our CollageGAN model against that of OASIS and found users preferred our approach 83.6\% times. Please note that for a fair comparison we did not use any additional training data for the CollageGAN model in this case.

 \begin{table}[t!]
 \scriptsize
     \begin{center}
     \resizebox{0.5\textwidth}{!}{
         \begin{tabular}{ c|c|c|c|c|c|c|c|c|c } 
             scenes & \multicolumn{5}{c}{Bedroom} & \multicolumn{2}{|c}{Cityscapes} & \multicolumn{2}{|c}{Human}\\
             \hline
             classes & chest & chair & pillow & lamp & table & car & person & face & shoe \\
             \hline
             FID (base) & 146.15 & 165.24 & 127.67 & 88.20  & 125.48 & 44.50 & 98.88 & 15.71 & 33.33 \\
             \hline
             FID (CollageGAN) & 132.12 & 155.52 & 136.79 & 80.12 &119.44 & 30.42 & 82.34 & 13.54 & 29.87 \\
             \hline
             User Favor & 71\%  & 70\% & 33\%  & 62\% & 60\% & 94\% & 89\% & 84\%& 69\% \\
            
         \end{tabular}}

 	    \caption{ \textbf{Per class FID and user preference.} For most classes, our CollageGAN model is preferred over our base model.}
 	    \label{table:perclass_res}

	\end{center}
	\vspace{-0.3in}
 \end{table}

 \begin{table}[t]
 
  \centering
  \resizebox{0.45\textwidth}{!}{
  \begin{tabular}{cccc|cc}
    \multirow{2}{*}{} &
      \multicolumn{3}{c}{FID $\downarrow$} &
      \multicolumn{2}{c}{User study $\uparrow$} \\
    & I: Base & II: w/o extra & III: w/ extra & I vs. II & I vs. III \\
    \midrule
    car & 44.50 & 36.71 & 30.42 & 23\%~/~77\% & 6\%~/~94\% \\
    person & 98.88 & 88.47 & 82.34 & 13\%~/~87\% & 11\%~/~89\% \\
    \bottomrule
  \end{tabular}}
  \caption{\textbf{Additional data ablation study.} I is our base model. II and III are CollageGAN models without and with extra data. }
  	\label{table:data_ablation}
  	\vspace{-0.2in}
\end{table}

\begin{figure*}[t!]
    \centering
    \includegraphics[width=0.98\textwidth]{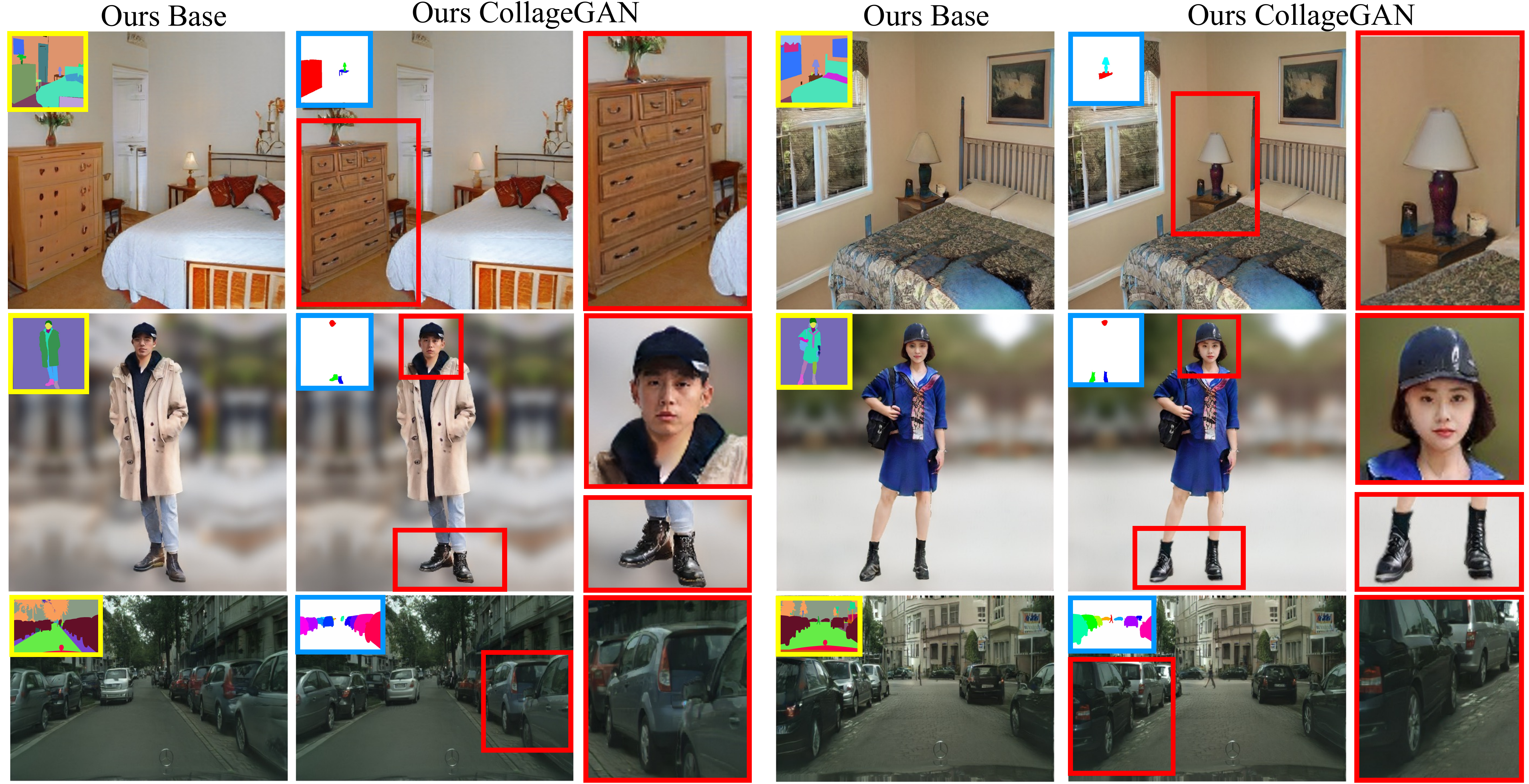}
    \caption{ \textbf{Comparing our base model with our CollageGAN model.} Our CollageGAN model takes advantage of class-specific generators to provide more details to specific classes in the image generated by the base model.}
    \label{fig:class_comparison}
    \vspace{-0.1in}
\end{figure*}

\vspace{0.1in}

\noindent\textbf{Qualitative results.} In Figure~\ref{fig:baselne_comparison}, we compare our base model against SPADE and OASIS. Our generated images look more realistic and detailed. For example, the bed sheet in generated bedrooms contain more textures and clothes on generated humans contain more wrinkles. We also found OASIS to have some boundary artifacts which can be seen by zooming in on the OASIS generated human images. Our model can also generate multiple images corresponding to the same segmentation map by sampling different $z$. We show these results in the supp.

\subsection{CollageGAN model results}
Our CollageGAN model consisting of class-specific generators can further improve the quality of the images generated by our base model which we evaluate next.

\vspace{0.5em}
\noindent\textbf{Quantitative results and human evaluation.}
We first compute per-class FID comparing our base model with our class-specific generators. Specifically, we crop each instance from both the original base image and the composition image generated using CollageGAN, and resize them to the average crop size over all instances in the class. 
We also conduct AMT evaluation, in which workers are shown class instances generated by our base model and by class-specific generator, and asked to choose the more realistic one. We show segmentation map at top with target instance highlighted, and two images at bottom (cropped from base and composition image). 
For each class, we show 100 pairs of images and 5 unique users are asked to judge each pair. In Table~\ref{table:perclass_res}, we report per-class FID of our base model (first row) and our CollageGAN model (second row) and also the percentage of time users prefer our class-specific generator over the base model. Our class-specific generator consistently obtains lower FID and higher user preference in all object categories except for the pillow class. We observed that for the pillow class, the base model usually generates a pure white color pillow (which is the most common color in the training data) which is simple and usually looks realistic. Hence, our CollageGAN model does not add much value and any small boundary artifacts due to composition can become more noticeable.

\vspace{0.5em}
\noindent\textbf{Qualitative results.}In Figure~\ref{fig:class_comparison}, we show results of compositing the pixels generated from our class-specific generator on top of the image generated by the base model. In the first column, we show the segmentation mask (yellow inset) and the corresponding image generated by our base generator. In the second column, we show our CollageGAN model results (highlighted class instance in blue inset show composited class instances). Finally, in the third column, we show the zoomed in view of the composited class instances which contain more details. For example, the chest in the first row contains more structural details like box boundaries and handles. The face region in the second row and the car in the last row look more realistic.

\begin{figure}[t!]
    \centering
    \includegraphics[width=0.4\textwidth]{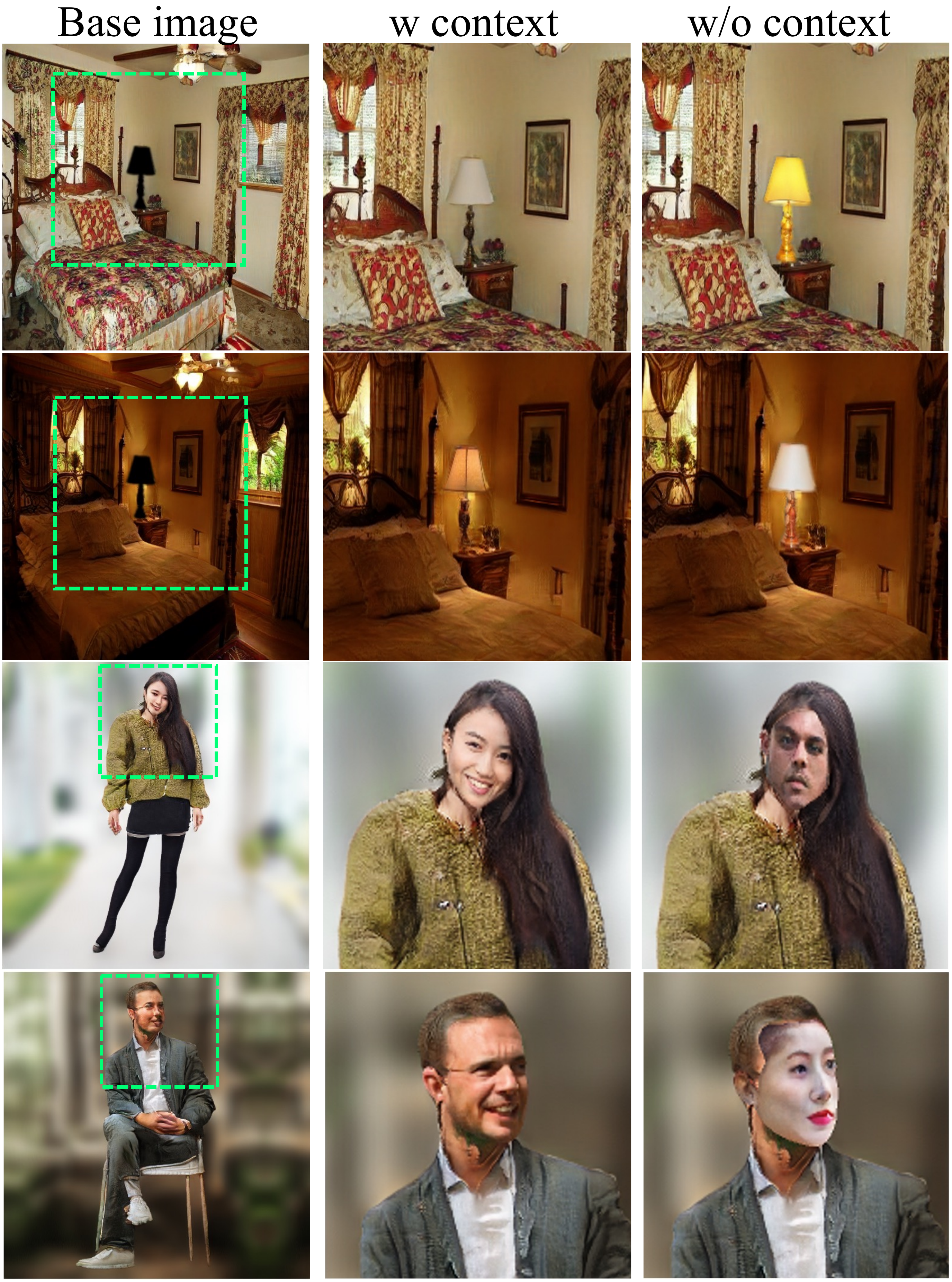}
    \caption{\textbf{Context analysis.} The last two columns show the importance of context during our class-specific model training.
    }
    \label{fig:study_context}
    \vspace{-0.05in}
\end{figure}

\vspace{0.5em}
\noindent\textbf{Ablation study.} We first study the impact of training our class-specific generators with additional training data.  Table~\ref{table:data_ablation} reports the result on cityscapes dataset (refer supp for the bedroom dataset), we can see that our class-specific generators perform better than the base model even without using any additional data (both in terms of user preference and FID). This shows the advantage of class-specific weights and centrally aligned data. Using additional training data further improves FID and user preference.

Next, we study the impact of providing context information $C$ as input to our class-specific generators. In Fig.~\ref{fig:study_context}, our lamp generator trained with context generates lamps that are consistent with the surrounding lighting condition whereas the model trained without context fails to do so. For this ablation study, our lamp generator trained with context does not use blurred foreground during the training and inference time, so the network only relies on context to determine the lamp color. Similarly, the last two rows indicate that the model may fail to infer correct gender and skin color when trained without any context. We also observe that sometimes our class-specific generator trained with context can correct the mistakes (like incorrect car orientations in Figure~\ref{fig:class_comparison} bottom row) made by our base generator. Please see the supp for more results.

\begin{figure}[t!]
    \centering
    \includegraphics[width=0.45\textwidth]{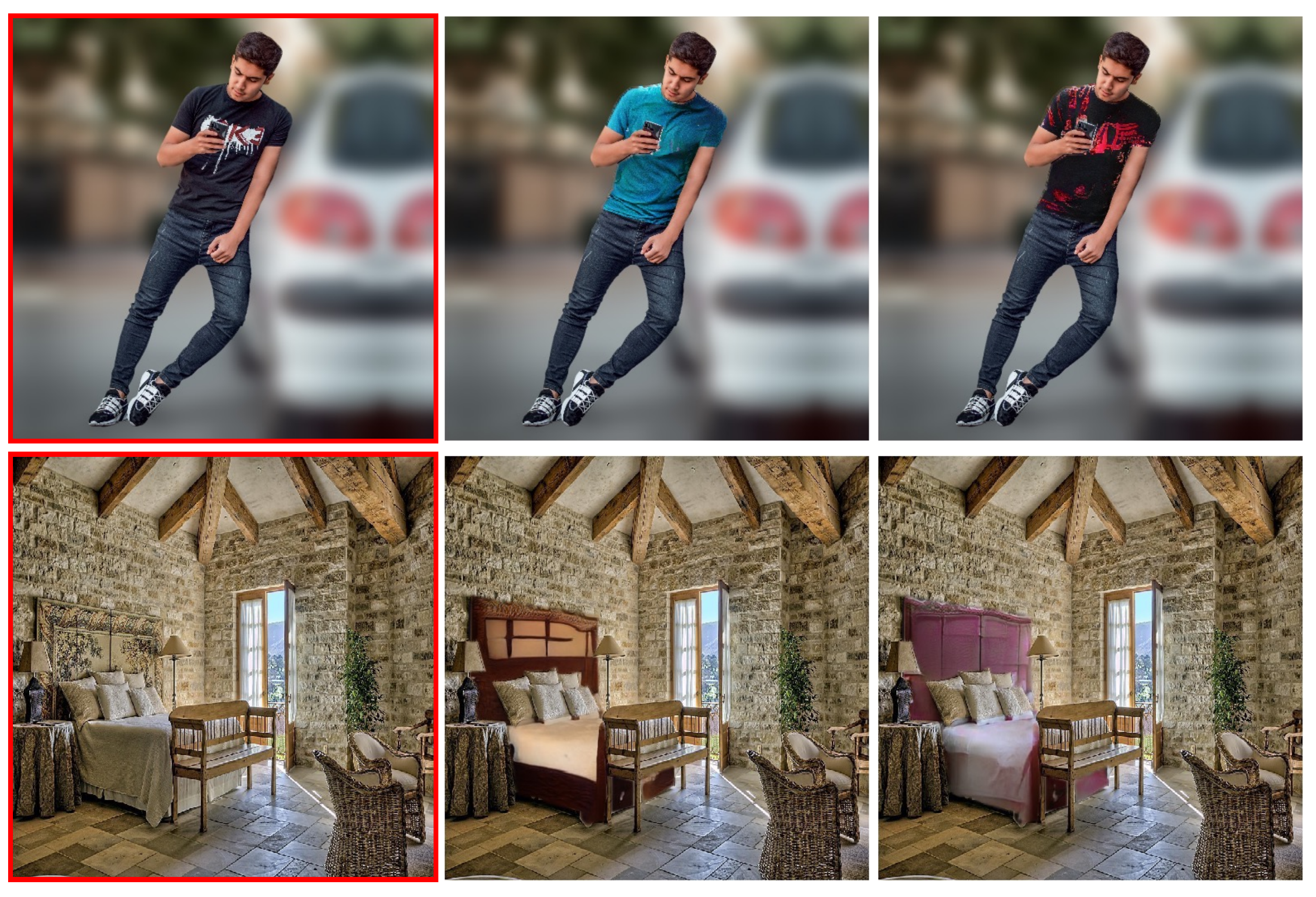}
    \caption{ \textbf{Replacing class instances in the real image}. Images in the red box are real images. Here bed and uppercloth generator is used to replace the original objects. }
    \label{fig:replace_real}
    \vspace{-0.1in}
\end{figure}

\vspace{0.5em}
\noindent\textbf{Applications.} Apart from generating more detailed and realistic images, class-specific generators can be used for several interesting applications. In Fig.~\ref{fig:replace_real}, we use our class-specific generators to replace a single object in the real image without affecting the other image parts. We mask out the object instance we want to replace and give the remaining real image as context for our class-specific generator. In Fig.~\ref{fig:replace_real}, we replace the shirt in the top row and the bed in the bottom row. Our generated region looks harmonious with the rest of the image while being different from the original. Note that here we train new bed and upper clothes generators by masking out foreground region as we want the model to generate diverse textures rather than following the original color tone in the real image.

\begin{figure}[t!]
    \centering
    \includegraphics[width=0.45\textwidth]{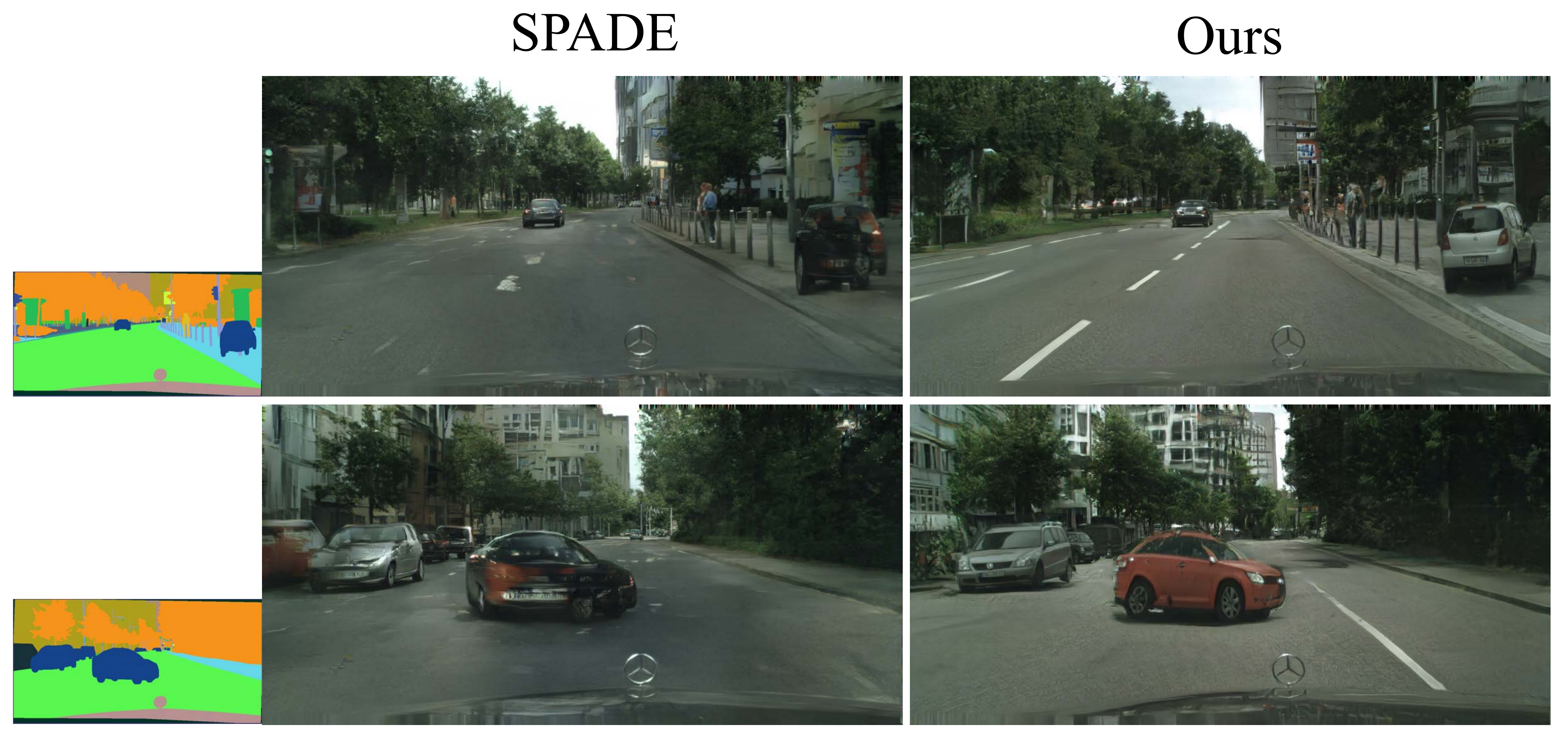}
    \caption{ \textbf{Out-of-distribution generation.} Our car specific generator can generate cars on the sidewalk or on the middle of the road, which do not exist in the training distribution.}    
    \label{fig:ood}
    \vspace{-0.1in}
\end{figure}

Both our base model and SPADE-based models are trained to generate the whole scene at once and as a result it is difficult for them to generate objects outside their original context. In contrast, our class-specific generator can achieve this as they are trained to only generate a fixed class. Fig.~\ref{fig:ood}, shows that our class-specific generator can generate a car on sidewalks instead of road while SPADE fails. Finally, we can also use our class-specific generator to generate a very high resolution image ($4096 \times 4096$) where important parts like face can be generated at high resolution while the remaining parts can still be generated at lower resolution by our base model. Please see supp for these results.  

\vspace{-0.05in}
\section{Discussion and Limitations}
\vspace{-0.05in}
We introduced a conditional version of StyleGAN2 architecture as a powerful base model for high quality and high resolution semantic image generation. We used the idea of leveraging context-aware class-specific generators to further enhance the quality of results, especially for small foreground objects. Our base model generates more realistic images than baselines, but does not perfectly align with the input segmentation map. The reason is that we downsample the segmentation map to $32 \times 32$ resolution (feature tensor $\phi\prime$) before injecting it into the decoder. We argue that weak conditioning from the segmentation map could give our model more freedom to generate realistic pixels and make our results more robust to inaccurate segmentation maps due to human or algorithm mistakes. In the supp, we provide a detailed study to evaluate the alignment between the input segmentation map and the generated images for our base model and baseline approaches.

\vspace{-10pt}
\paragraph{Acknowledgements.} This work was supported in part by
NSF CAREER IIS-1751206.

{\small
\bibliographystyle{ieee_fullname}
\bibliography{refs}

\begin{thebibliography}{10}\itemsep=-1pt

\bibitem{arjovsky-icml2017}
Martin Arjovsky, Soumith Chintala, and L{\'e}on Bottou.
\newblock Wasserstein gan.
\newblock {\em ICML}, 2017.

\bibitem{brock-iclr2019}
Andrew Brock, Jeff Donahue, and Karen Simonyan.
\newblock Large scale {GAN} training for high fidelity natural image synthesis.
\newblock In {\em ICLR}, 2019.

\bibitem{chen-nips16}
Xi Chen, Yan Duan, Rein Houthooft, John Schulman, Ilya Sutskever, and Pieter
  Abbeel.
\newblock Infogan: Interpretable representation learning by information
  maximizing generative adversarial nets.
\newblock In {\em NeurIPS}, 2016.

\bibitem{cordts-cvpr2016}
Marius Cordts, Mohamed Omran, Sebastian Ramos, Timo Rehfeld, Markus Enzweiler,
  Rodrigo Benenson, Uwe Franke, Stefan Roth, and Bernt Schiele.
\newblock The cityscapes dataset for semantic urban scene understanding.
\newblock In {\em CVPR}, 2016.

\bibitem{cui-cvpr2019}
Yin Cui, Menglin Jia, Tsung-Yi Lin, Yang Song, and Serge Belongie.
\newblock Class-balanced loss based on effective number of samples.
\newblock 2019.

\bibitem{imaterialist}
CVPR-FGVC5.
\newblock https://www.kaggle.com/c/imaterialist-challenge-furniture-2018.
\newblock 2018.

\bibitem{Kingma-iclr2014}
Max~Welling Diederik P~Kingma.
\newblock Auto-encoding variational bayes.
\newblock In {\em ICLR}, 2014.

\bibitem{Dollar-cvpr2009}
Piotr Dollar, Christian Wojek, Bernt Schiele, and Pietro Perona.
\newblock Pedestrian detection: A benchmark.
\newblock In {\em CVPR}, 2009.

\bibitem{goodfellow-gantutorial2016}
Ian Goodfellow.
\newblock {NeurIPS} 2016 tutorial: Generative adversarial networks.
\newblock {\em arXiv preprint arXiv:1701.00160}, 2016.

\bibitem{goodfellow-nips2014}
Ian Goodfellow, Jean Pouget-Abadie, Mehdi Mirza, Bing Xu, David Warde-Farley,
  Sherjil Ozair, Aaron Courville, and Yoshua Bengio.
\newblock Generative adversarial nets.
\newblock In {\em NeurIPS}, 2014.

\bibitem{gu-cvpr2019}
Shuyang Gu, Jianmin Bao, Hao Yang, Dong Chen, Fang Wen, and Lu Yuan.
\newblock Mask-guided portrait editing with conditional gans.
\newblock In {\em CVPR}, 2019.

\bibitem{gulrajani-nips17}
Ishaan Gulrajani, Faruk Ahmed, Martin Arjovsky, Vincent Dumoulin, and Aaron~C
  Courville.
\newblock Improved training of wasserstein gans.
\newblock In {\em NeurIPS}, 2017.

\bibitem{he-cvpr2016}
Kaiming He, Xiangyu Zhang, Shaoqing Ren, and Jian Sun.
\newblock Deep residual learning for image recognition.
\newblock 2016.

\bibitem{fid}
Martin Heusel, Hubert Ramsauer, Thomas Unterthiner, Bernhard Nessler, Gunter
  Klambauer, and Sepp Hochreiter.
\newblock Gans trained by a two time-scale update rule converge to a local nash
  equilibrium.
\newblock In {\em NeurIPS}, 2017.

\bibitem{hinz-iclr019}
Tobias Hinz, Stefan Heinrich, and Stefan Wermter.
\newblock Generating multiple objects at spatially distinct locations.
\newblock In {\em ICLR}, 2019.

\bibitem{huang-iccv2017}
Rui Huang, Shu Zhang, Tianyu Li, and Ran He.
\newblock Beyond face rotation: Global and local perception gan for
  photorealistic and identity preserving frontal view synthesis.
\newblock In {\em ICCV}, 2017.

\bibitem{isola-pix2pix2017}
Phillip Isola, Jun-Yan Zhu, Tinghui Zhou, and Alexei Efros.
\newblock Image-to-image translation with conditional adversarial networks.
\newblock In {\em CVPR}, 2017.

\bibitem{johnson-eccv2016}
Justin Johnson, Alexandre Alahi, and Li Fei-Fei.
\newblock Perceptual losses for real-time style transfer and super-resolution.
\newblock In {\em ECCV}, 2016.

\bibitem{Johnson-cvpr2018}
Justin Johnson, Agrim Gupta, and Li Fei-Fei.
\newblock Image generation from scene graphs.
\newblock In {\em CVPR}, 2018.

\bibitem{karras-iclr2018}
Tero Karras, Timo Aila, Samuli Laine, and Jaakko Lehtinen.
\newblock Progressive growing of gans for improved quality, stability, and
  variation.
\newblock {\em ICLR}, 2018.

\bibitem{karras-cvpr2019}
Tero Karras, Samuli Laine, and Timo Aila.
\newblock A style-based generator architecture for generative adversarial
  networks.
\newblock In {\em CVPR}, 2019.

\bibitem{karras-cvpr2020}
Tero Karras, Samuli Laine, Miika Aittala, Janne Hellsten, Jaakko Lehtinen, and
  Timo Aila.
\newblock Analyzing and improving the image quality of stylegan.
\newblock In {\em CVPR}, 2020.

\bibitem{tuoma-nips2019}
Tuomas Kynkäänniemi, Tero Karras, Samuli Laine, Jaakko Lehtinen, and Timo
  Aila.
\newblock Improved precision and recall metric for assessing generative models.
\newblock In {\em NeurIPS}, 2019.

\bibitem{lee-cvpr2020}
Cheng-Han Lee, Ziwei Liu, Lingyun Wu, and Ping Luo.
\newblock Maskgan: Towards diverse and interactive facial image manipulation.
\newblock In {\em IEEE Conference on Computer Vision and Pattern Recognition
  (CVPR)}, 2020.

\bibitem{lewis-arxiv2020}
Kathleen~M Lewis, Srivatsan Varadharajan, and Ira Kemelmacher-Shlizerman.
\newblock Vogue: Try-on by stylegan interpolation optimization.
\newblock {\em arXiv preprint arXiv:2101.02285}, 2021.

\bibitem{li-cvpr2018}
Peipei Li, Yibo Hu, Qi Li, Ran He, and Zhenan Sun.
\newblock Global and local consistent age generative adversarial networks.
\newblock In {\em CVPR}, 2018.

\bibitem{lin-cvpr2017}
Tsung-Yi Lin, Piotr Dollár, Ross Girshick, Kaiming He, Bharath Hariharan, and
  Serge Belongie.
\newblock Feature pyramid networks for object detection.
\newblock In {\em CVPR}, 2017.

\bibitem{liu-nips2019}
Xihui Liu, Guojun Yin, Jing Shao, Xiaogang Wang, and Hongsheng Li.
\newblock Learning to predict layout-to-image conditional convolutions for
  semantic image synthesis.
\newblock In {\em NeurIPS}, 2019.

\bibitem{Miyato-iclr2018}
Takeru Miyato, Toshiki Kataoka, Masanori Koyama, and Yuichi Yoshida.
\newblock Spectral normalization for generative adversarial networks.
\newblock In {\em ICLR}, 2018.

\bibitem{park-cvpr2019}
Taesung Park, Ming{-}Yu Liu, Ting{-}Chun Wang, and Jun{-}Yan Zhu.
\newblock Semantic image synthesis with spatially-adaptive normalization.
\newblock In {\em CVPR}, 2019.

\bibitem{sarkar-arxiv2021}
Kripasindhu Sarkar, Vladislav Golyanik, Lingjie Liu, and Christian Theobalt.
\newblock Style and pose control for image synthesis of humans from a single
  monocular view.
\newblock In {\em ArXiv:2102.11263}, 2021.

\bibitem{simonyan-iclr2015}
K. Simonyan and A. Zisserman.
\newblock {Very Deep Convolutional Networks for Large-Scale Image Recognition}.
\newblock In {\em ICLR}, 2015.

\bibitem{singh-cvpr2019}
Krishna~Kumar Singh, Utkarsh Ojha, and Yong~Jae Lee.
\newblock {FineGAN}: Unsupervised hierarchical disentanglement for fine-grained
  object generation and discovery.
\newblock In {\em CVPR}, 2019.

\bibitem{Sushko-iclr2021}
Vadim Sushko, Edgar Schönfeld, Dan Zhang, Juergen Gall, Bernt Schiele, and
  Anna Khoreva.
\newblock You only need adversarial supervision for semantic image synthesis.
\newblock In {\em ICLR}, 2021.

\bibitem{tang-cvpr2020}
Hao Tang, Dan Xu, Yan Yan, Philip H.~S. Torr, and Nicu Sebe.
\newblock Local class-specific and global image-level generative adversarial
  networks for semantic-guided scene generation.
\newblock In {\em CVPR}, 2020.

\bibitem{wang-cvpr2018}
Ting-Chun Wang, Ming-Yu Liu, Jun-Yan Zhu, Andrew Tao, Jan Kautz, and Bryan
  Catanzaro.
\newblock High-resolution image synthesis and semantic manipulation with
  conditional gans.
\newblock In {\em CVPR}, 2018.

\bibitem{wu-github2019}
Yuxin Wu, Alexander Kirillov, Francisco Massa, Wan-Yen Lo, and Ross Girshick.
\newblock Detectron2.
\newblock \url{https://github.com/facebookresearch/detectron2}, 2019.

\bibitem{yang-iclr17}
Jianwei Yang, Anitha Kannan, Dhruv Batra, and Devi Parikh.
\newblock Lr-gan: Layered recursive generative adversarial networks for image
  generation.
\newblock {\em ICLR}, 2017.

\bibitem{zhang-iccv17}
Han Zhang, Tao Xu, Hongsheng Li, Shaoting Zhang, and Dimitris Metaxas.
\newblock Stackgan: Text to photo-realistic image synthesis with stacked
  generative adversarial networks.
\newblock In {\em ICCV}, 2017.

\bibitem{zhou-TPAMI2017}
Bolei Zhou, Agata Lapedriza, Aditya Khosla, Aude Oliva, and Antonio Torralba.
\newblock Places: A 10 million image database for scene recognition.
\newblock {\em PAMI}, 2017.

\bibitem{zhou-cvpr2017}
Bolei Zhou, Hang Zhao, Xavier Puig, Tete Xiao, Sanja Fidler, Adela Barriuso,
  and Antonio Torralba.
\newblock Scene parsing through ade20k dataset.
\newblock In {\em CVPR}, 2017.

\bibitem{zhu-iccv2017}
Jun{-}Yan Zhu, Taesung Park, Phillip Isola, and Alexei~A. Efros.
\newblock Unpaired image-to-image translation using cycle-consistent
  adversarial networks.
\newblock {\em ICCV}, 2017.

\bibitem{Zhu-cvpr2020}
Peihao Zhu, Rameen Abdal, Yipeng Qin, and Peter Wonka.
\newblock Sean: Image synthesis with semantic region-adaptive normalization.
\newblock In {\em CVPR}, 2020.

\end{thebibliography}
}

\clearpage
\section*{Appendix}

In this supplementary material, we first elaborate our network architecture and its ablation study. Next, we provide more training and implementation details. After that we discuss the alignment between the input segmentation map and the generated images and also discuss about the diversity of our model. Finally, we show more training data ablation study on bedroom dataset and provide more qualitative results.

\section{Architecture details and its ablation}

\begin{figure}[t!]
    \centering
    \includegraphics[width=0.45\textwidth]{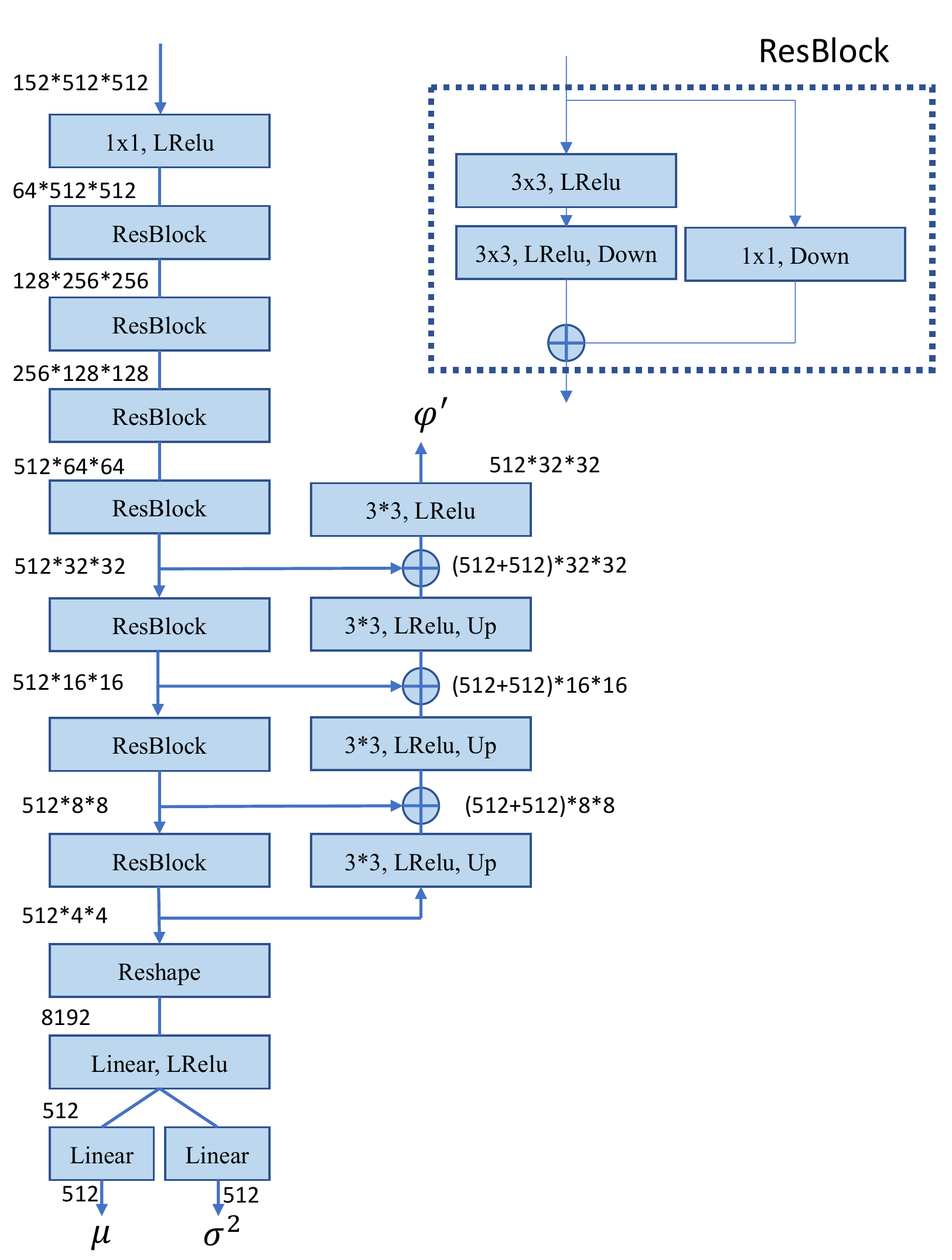}
    \caption{The encoder details. This includes both encoder1 and encoder2 in the Fig. 2 in the main paper. Up and Down stand for upsampling and downsampling operation.}    
    \label{fig:encoder}
    \vspace{-0.1in}
\end{figure}

We use our base model for the $512 \times 512$ resolution as an example to demonstrate our architecture. Fig.~\ref{fig:encoder} shows the encoder architecture which we use for the bedroom dataset, where we have 151 semantic labels and 1 edge map. The input is converted into a fixed 64-channel feature by the first $1 \times 1$ convolution. Then this feature will be passed into a series of ResBlock (details of the ResBlock are shown on the top right of Fig.~\ref{fig:encoder}). Once the feature resolution reaches $4\times4$, one branch will flatten it and calculate mean and variance through fully-connected layers. Another branch (named top-down pathway in the main paper) will process this feature with a few layers of convolution and upsampling. In the meantime, the top-down pathway will aggregate previous features with lateral connection to preserve better spatial alignment. Eventually this pathway will output $\phi\prime$ which is fed into the feature cropping module as shown in the Fig.2 in the main paper. We set resolution of $\phi\prime$ 16 times smaller than the input in all our experiments.

Note that a non-square input can be also given to generate non-square shape $\phi\prime$. For example, in the Cityscapes dataset, the input has the resolution of $512 \times 1024$, then this encoder will yield $\phi\prime$ with the size of $32 \times 64$. A small modification is needed for the first linear layer whose input should be $512 \times 4 \times 8$ in this case. The encoder architecture for the other cases, including our class-specific models, is similar with adding/removing the ResBlock depending on the input resolution. 

\begin{figure}[t!]
    \centering
    \includegraphics[width=0.45\textwidth]{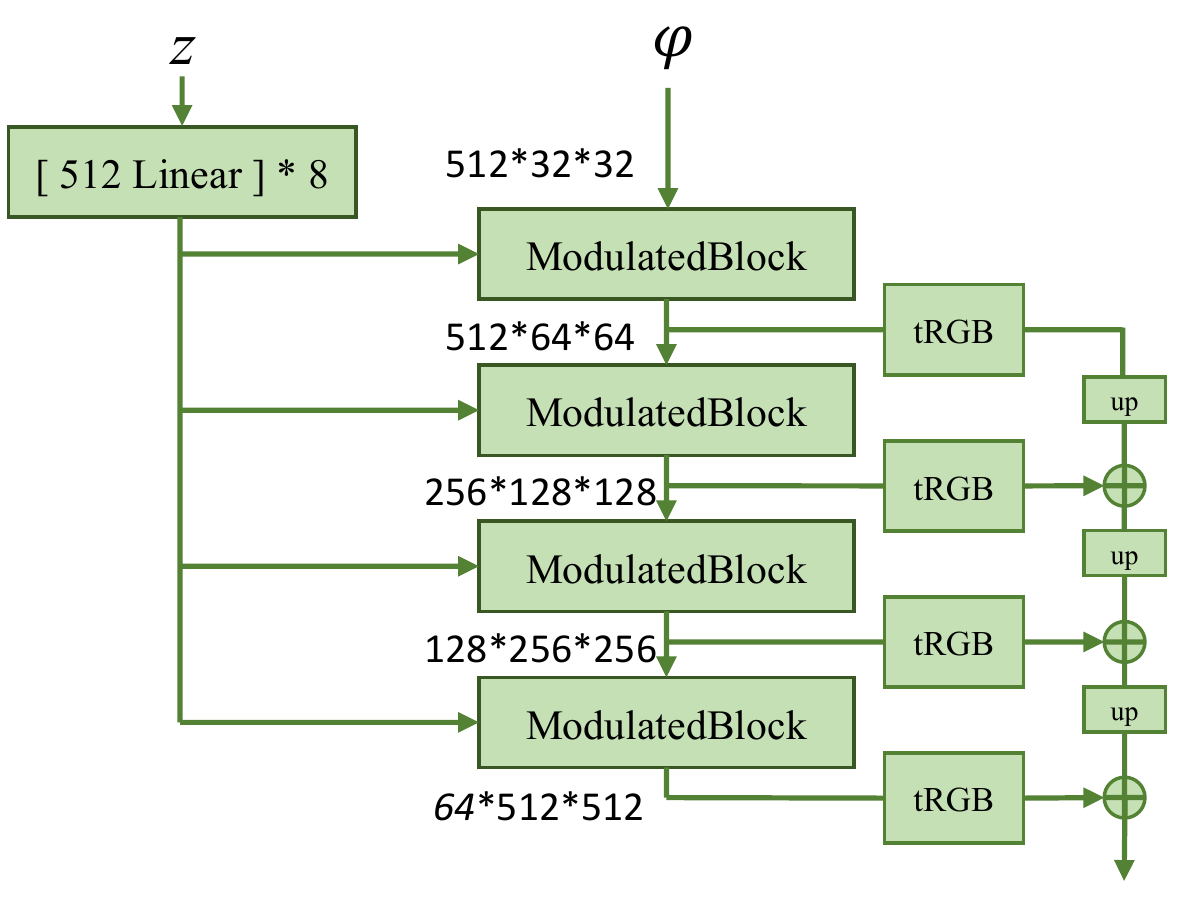}
    \caption{The decoder details. We adopt it from StyleGAN2 but without the constant input.}    
    \label{fig:decoder}
    \vspace{-0.1in}
\end{figure}

The decoder architecture is same with that of StyleGAN2~\cite{karras-cvpr2020} with one exception that the input is our starting feature tensor $\phi$ instead of learnable constant as shown in Fig.~\ref{fig:decoder}. Our discriminator is same as the one used in~\cite{karras-cvpr2020}. Please refer~\cite{karras-cvpr2020} for more details. 

\begin{figure*}[t!]
    \centering
    \includegraphics[width=0.92\textwidth]{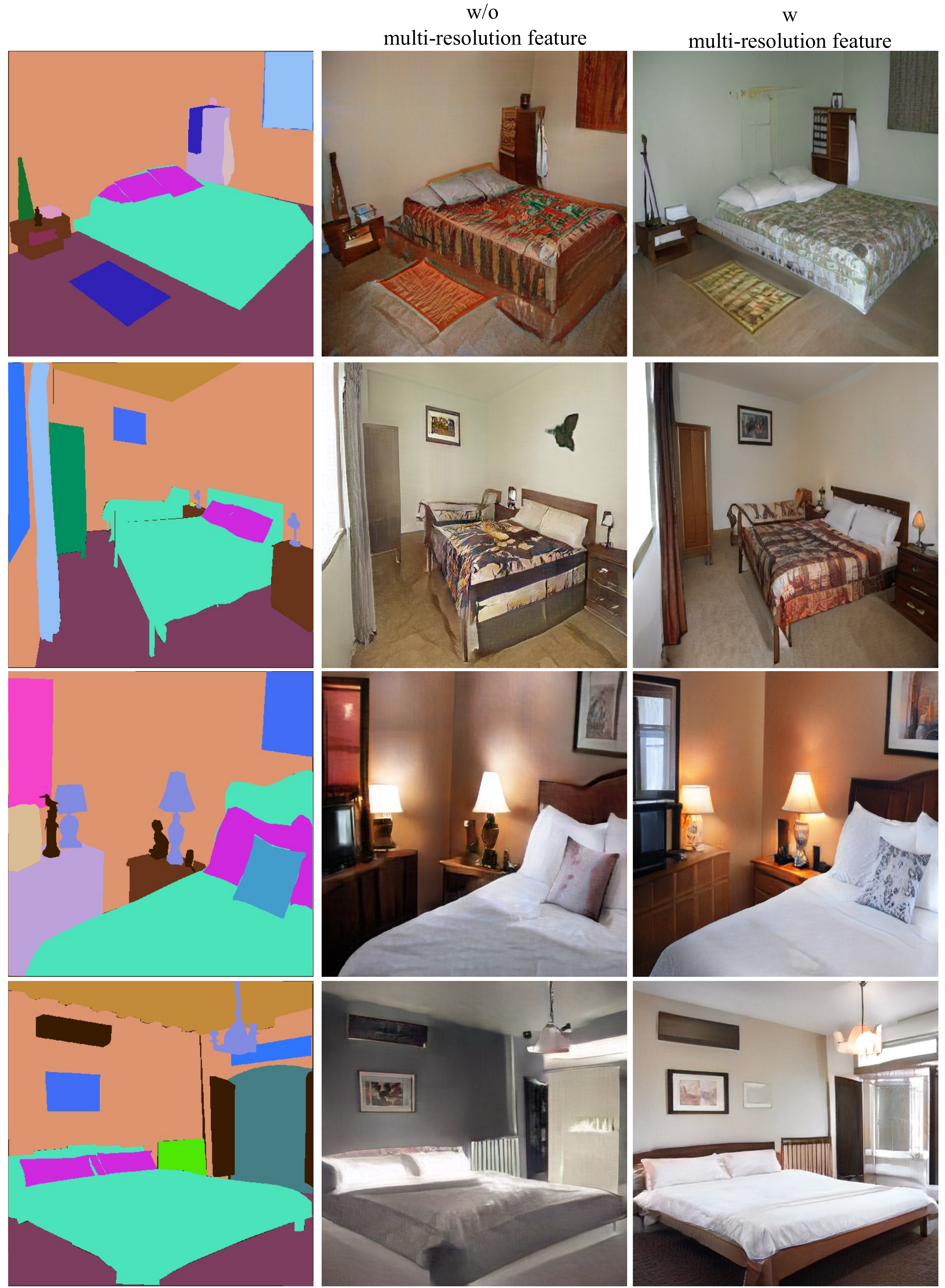}
    \caption{ The middle column is the result of architecture without multi-resolution feature aggregation. We can see that using multi resolution features add more details to the image. For example, chest in first three rows and bed in last row have more details with multi resolution features. }    
    \label{fig:ablation}
    \vspace{-0.1in}
\end{figure*}

As mentioned in the main paper, our encoder designed in this way, so that it can provide the decoder merged multi-resolution features $\phi$. The higher resolution features are more accurately localized to the input where as lower resolution features are semantically stronger and have more global information. We conduct an ablation study to show the importance of multi-resolution features. Specifically, we remove our top-down path way in the encoder and directly use the output of encoder1 (the $512\times32\times32$ feature outputted from the 4th ResBlock in the Fig.~\ref{fig:encoder} ) as the starting feature for the decoder. We found that the image quality is not as good as our base model. The qualitative result is shown in Fig.~\ref{fig:ablation}. The FID score 40.88 which is higher than our base model result 34.41 (Table 2 in the main paper) is consistent with the visual quality comparison.

\section{Datasets details}

Here we provide more details about our extra datasets used by our class-specific models. In order to be consistent, we use the same dataset index as used the main paper. 

(4) \textbf{iMaterialist}~\cite{imaterialist}.  It contains images of 128 categories of furniture. We selected 36 categories commonly appearing in the bedroom and use the segmentor~\cite{wu-github2019} trained on ADE20K to get the masks. Totally, we have 50,370 images.

(5) \textbf{Indoor}. We selected $childs\_room$,  $dining\_room$ and $living\_room$ from the Places dataset~\cite{zhou-TPAMI2017} and apply the segmentor trained on ADE20K to get masks. Note that we use $bedroom$ and $hotel\_room$ from the Places~\cite{zhou-TPAMI2017} to train our base model, whereas these three categories are used for class-specific generator training.

(6) \textbf{Cityscapes\_extra}~\cite{cordts-cvpr2016}. Except for commonly used 3,000 images with annotations, there are also extra 19,998 training images officially provided. We train a segmentor~\cite{wu-github2019} to get masks for those images. 

(7) \textbf{Caltech Pedestrian}~\cite{Dollar-cvpr2009}. 10 hours of video taken from a vehicle driving through urban environment. We extract every alternate frames to get 124,942 images and we use the segmentation model trained on the Cityscapes to obtain pseudo labels.

(8) \textbf{CelebaMask-HD}~\cite{lee-cvpr2020}. It includes 30,000 segmentation masks for the CelebAHQ face image dataset. There are 19 different region categories. To be compatible with our full human body dataset annotation, we merge some categories such as eyes, mouth, skin, nose into face category. We only use this dataset to demonstrate our mixed-resolution application in Fig.~\ref{fig:mixed1}$-$\ref{fig:mixed2} as this dataset can provide us very high resolution face images.

\section{Training details}

For our class-specific generator, the following classes are trained at $128 \times 128$ resolution: person (Cityscapes), shoes (Human) and face (Human). For the other classes in the main paper, we train them at $256 \times 256$ resolution. We choose the resolution for each class based on their average size in training data. Please note that the average training data size is not from base image size, but from the raw training data size. For example, if we choose face from $512 \times 512$ real full human body images, which are used to train base model, then it is difficult to get $128 \times 128$ face images since the face region only occupies a small portion of the image. However, since the raw image we collected is at least at 1K resolution, we can acquire a lot of higher quality face images at $128 \times 128$ resolution from raw data. This is the advantage of having class-specific generator. To maintain the training data quality, we do not use cropped instance smaller than $64 \times 64$  and $128 \times 128$ for $128 \times 128$ and  $256 \times 256$ class-specific model training, respectively. How to fully take advantage of all training data could be an interesting future work.     

Since our discriminator expects a square shape input, for the Cityscapes dataset, we split each image ($1024 \times 512$) into two $512 \times 512$ images and stack them together before feeding into the discriminator. We have explored this by trying zero padding as an alternative way. Specifically, we pad the image into $1024 \times 1024$ before feeding into discriminator. However, we found this leads to worse results than splitting. The FID score of padding is 51.44 compared with 47.04 using splitting (Table2 in the main paper).   

We train our base model for 300K iterations with the batch size of 16 for bedroom and human dataset. For the Cityscapes, we only train 60K iterations with the same batch size due to less training data. For bedroom and human models, each of them only take 2 32GB V100 GPUs. Due to higher resolution for the Cityscapes ($1024 \times 512$), we use 4 V100 GPUs to train. However, based on our observation, it usuallly takes ~22GB memory per GPU, thus we hypothesize 4 GPUs with 24GB memory may also work. Our model is more friendly for training compared with SPADE-like architecture where we found it needs 8 32GB GPUs to train with the batch size of 16 at $512 \times 512$ resolution.  Among baselines we tried, we found LGGAN~\cite{tang-cvpr2020} consumes the most resources. We can not fit batch size of 1 in a single GPU when scale to higher resolution using their default code. Although for the cityscapes we were able to fit batch size of 1 by cutting their default --ngf hyperparameters from 32 to 28. (ngf controls numbers of channels, for example 32 channels becomes 28, 64 becomes 56).  We usually train 150K iterations for our class-specific models with the same batch size. Since we only train them at either 128 or 256 resolution, 1 32GB V100 is sufficient for each class. Note that since our class-specific model is not dependent on each other, thus 8 different classes can be trained in parallel in a 8-GPU machine.

We set the loss weight of KL-Divergence and perceptual loss as 0.01 and 1, respectively ($\lambda_1$ and $\lambda_2$ in the main paper ). Following the StyleGAN2~\cite{karras-cvpr2020}, we only conduct path regularization and r1 regularization every 4 and 16 iterations. The loss weight of these two terms are 2 and 10 (note that these two terms are not explicitly symbolized in the main paper, but there are within $\mathcal{L}_{stylegan}$). Also, the perceptual loss is applied every 4 iterations.

\section{Composition}

During the inference time, we need to composite instances generated by our class-specific generators. First, we need to crop our base image using enlarged bounding box of one instance to get its surrounding pixels $C_i$, and then either mask or blur out the instance according to the procedure used during training. Similarly, a cropped semantic mask $C_s$ is also acquired. The class-specific model takes in these two and generates an instance $I_c$.

Next, we will composite this new instance $I_c$ into the base image $I_b$. Since we know the original size and location of this instance in the base image, we will first create an alpha mask of this instance using ground truth instance mask $Ins$,
\begin{equation}
    M_{alpha}=
    \begin{cases}
      1, & \text{if}\ Ins(i,j)= target\_instance\_idx \\
      0, & \text{otherwise}
    \end{cases}
  \end{equation}
where $Ins$ is a 2D map with different values at each location, and each value is the index for a unique instance. The $target\_instance\_idx$ is the index for the current target instance. Then we will resize and relocate the generated instance $I_c$ into the correct position according to the $M_{alpha}$ to get the relocated generated instance $I_{c\_relocation}$. In order to avoid the potential small gaps due to quantization during the process of resizing and relocating, we will further dilate boundaries of both $M_{alpha}$ and $I_{c\_relocation}$. Finally the composition image $I_{comp}$ is 
\begin{equation}
    I_{comp} = {M'}_{alpha} \times {I'}_{c\_relocation} +  ( 1 - {M'}_{alpha} )\times {I_b},
\end{equation}
where ${M'}_{alpha}$ and ${I'}_{c\_relocation}$ are dilated ${M}_{alpha}$ and ${I}_{c\_relocation}$. After the composition is done for the first instance, $I_{comp}$ will be served as base image $I_{b}$ for the next instance.

\section{Alignment study}

As mentioned in Section 5 of the main paper, since we only provide the feature with $32 \times 32$ resolution (in the case of $512 \times 512$ base image generation) as the input to the decoder, our base image does not perfectly follow the input segmentation map at pixel level alignment. However, we argue that this could be a desired property sometimes. For example, in the application of generating an image based on the semantic map, users may not provide a perfect semantic map and a model with flexibility to alter boundaries of objects may have potential to generate more realistic objects. Nevertheless, we still conducted a series of studies to study this problem.

\begin{table}[t!]
    \begin{center}
        \scriptsize
        \begin{tabular}{ c|c|c|c|c|c|c } 
            &  Bedroom & Human & City &  Bedroom(f) & Human(f) & City(f) \\
            \hline
            SPADE & 49.28 & 79.40 & 33.83 & 64.42 & 94.02 & 68.40\\
            \hline
            OASIS &   63.15 & 82.85 & 44.56  & 68.37 & 93.62 & 68.80\\
            \hline
            Ours & \textbf{64.59} & \textbf{83.30} & \textbf{45.01} &   \textbf{74.31} & \textbf{95.06} & \textbf{82.64} \\
        \end{tabular}
	    \caption{ \textbf{Object level alignment.} Top-1 classification accuracy in cropped generated instances in three different datasets for all classes (left three columns) and only foreground objects for which we have class-specific generators (right three columns)}
	    \label{table:object-alignment}
	\end{center}
\vspace{-0.3in}
\end{table}

In order to verify the alignment in human perception, we first conduct a user study. Specifically, we show segmentation
map at top and two images at bottom (base image and baseline) to AMT workers and ask them which image corresponds better to the semantic mask. We explicitly ask the workers to not evaluate the image quality. Please see the screen shot in Fig.~\ref{fig:user_correspondence}. Instead of forcing them to choose one image, we also give them the third option (i.e., ``similar'') for the cases that are hard to tell which one is better.  Table~\ref{table:alignment} shows the results. Surprisingly, we found that, according to study participants, our base image has better correspondence with respect to the input semantic map in bedroom dataset. We attribute this to the fact that we generate much better quality images compared with baselines (please refer to Table 3 in the main paper). Some poorly synthesized object instances may not be easily recognizable in the baseline results, thus leading to worse perceptual alignment quality compared with ours. 

On the human dataset, baselines perform better than our approach for the alignment evaluation. But we are better or equally good compared to baselines around half of the times. On the Cityscapes dataset, our base model has worse alignment quality compared with baselines. A potential solution to fix the alignment issue caused by our base model is to use CollageGAN. As our CollageGAN model uses the ground-truth instance mask while training class-specific models and then performs composition using the instance mask, the final generated image would be better aligned to the segmentation mask. Thus we did another user study to compare our CollageGAN results with OASIS~\cite{Sushko-iclr2021} on the Cityscapes dataset, where our base model is inferior. The result is shown in the bracket in the Table~\ref{table:alignment}. We observe that after the composition, users prefer our results more. Also, our class-specific model is not dependent on our base model.
Thus, if one wants to have a better-aligned base image, they can choose a baseline approach such as OASIS to generate the base image and then use our class-specific generator idea to enhance different local details.   

As mentioned earlier, since our model may not generate images which are perfectly pixel-wise aligned with inputs, thus we conduct a quantitative evaluation by comparing our CollageGAN model and baselines at object-level alignment. Specifically, we first crop each instance for all classes in real images and train an object classification model using Resnet50~\cite{he-cvpr2016}. We train three different classifiers for three different datasets (bedroom, cityscapes and human), respectively. In order to tackle data imbalance issue we adopted the technique proposed in~\cite{cui-cvpr2019}. In the Table~\ref{table:object-alignment}, we report top-1 classification accuracy for three different approaches. The first three columns indicate we have best results, reflecting that our model generates better image quality. The last three columns indicate that having class-specific generators can further improve the results.

\section{Diversity}
To understand how diverse our generated images can be, we also quantitatively evaluate diversity of our model. We compare our base model with baselines. Here we report recall ($\uparrow$) to directly measure diversity~\cite{tuoma-nips2019}. The results for SPADE/OASIS/Ours on there datasets: Bedroom: 0.561/0.741/\textbf{0.776}; Human: 0.236/\textbf{0.776}/0.744; City: 0.342/0.724/\textbf{0.733}. The result indicates that our base model and OASIS has similar performance and they are both better than SPADE, which we hypothesize that it is due to the fact that the SPADE has a strong reconstruction supervision during training (high loss weight for perceptual loss~\cite{johnson-eccv2016} and feature match loss~\cite{wang-cvpr2018}), forcing the model to generate neutral color (This can be observed from the background in the human dataset results; Fig.~\ref{fig:baseline_human} ).

\section{User study interface}
Totally, we conducted three types of user study. The first one is correspondence study between our base model result and baseline result. The second one is realism study between our base model result and baseline result. The last one is realism study between our base model result and CollageGAN result. Their interfaces are shown in Fig.~\ref{fig:user_correspondence}$-$\ref{fig:user_instance}

\section{Additional data ablation study}

Since our class-specific model can be trained using data from other sources, thus, in this section, we study the importance of adding extra data.  Table~\ref{table:supp_data_ablation} shows results on the cityscapes dataset (top) and the bedroom dataset (bottom). For classes in the cityscapes, our class-specific generators perform better than the base model without additional data. By adding more data, the results will be even better in terms of both FID and user study. In the bedroom dataset, according to the FID, our class-specific model without extra data is doing better than the base model in the most cases, except for pillow class. For human evaluation, class-specific model without extra data is again better for all the classes except for lamp and pillow. For lamp, we end up removing lot of images as we want lamps to be bigger than $128 \times 128$ for training class specific generator. But our lamp model trained with additional data can easily outperform the base model in term of human evaluation. Also, we can see that adding more data improves the performance of all the classes consistently. This proves that one of advantages of having separate models is using additional data to boost the performance.

 \begin{table}[t]
 
  \centering
  \resizebox{0.45\textwidth}{!}{
  \begin{tabular}{cccc|cc}
    \multirow{2}{*}{} &
      \multicolumn{3}{c}{FID $\downarrow$} &
      \multicolumn{2}{c}{User study $\uparrow$} \\
    & I: Base & II: w/o extra & III: w/ extra & I vs. II & I vs. III \\
    \midrule
    car & 44.50 & 36.71 & 30.42 & 23\%~/~77\% & 6\%~/~94\% \\
    person & 98.88 & 88.47 & 82.34 & 13\%~/~87\% & 11\%~/~89\% \\
    \midrule 
    chest & 146.15 & 137.65 & 132.12 & 38\%~/~62\% & 29\%~/~71\% \\
    chair & 165.24 & 161.00 & 155.52 & 43\%~/~57\% & 30\%~/~70\% \\
    pillow & 127.67 & 135.58 & 136.79 & 70\%~/~30\% & 67\%~/~33\% \\
    lamp & 88.20 & 84.70 & 80.12 & 66\%~/~34\% & 38\%~/~62\% \\
    table & 125.48 & 116.39 & 119.44 & 41\%~/~59\% & 40\%~/~60\% \\
    \bottomrule
  \end{tabular}}
  \caption{\textbf{Additional data ablation study.} I is our base model. II and III are CollageGAN models without and with extra data. }
  	\label{table:supp_data_ablation}
  
\end{table}

\begin{table}[t!]
    \begin{center}
        \scriptsize
        \begin{tabular}{ c|c|c|c } 
            Datasets & Ours vs SPADE & Ours vs OASIS & Ours vs LGGAN \\
            \hline
            Bedroom & 66.8/4/29.6 & 59.6/4/36.4 & NA \\
            \hline
            Human & 34/22.4/43.6 & 31.6/15.2/53.2 & NA \\
            \hline
            Cityscapes & 17.2/13.2/69.6 & \vtop{\hbox{\strut 18.2/5.7/76.8 }\hbox{\strut (46.8/6.8/46.4)}}   & 21.2/13.6/65.2   \\
        \end{tabular}
	    \caption{The three numbers in each cell indicate percent of times our model is preferred vs similar vs baseline is preferred in terms of alignment. The CollageGAN result is in the bracket for Ours vs OASIS on cityscapes dataset.}
	    \label{table:alignment}
	\end{center}
\vspace{-0.3in}
\end{table}

\section{Additional qualitative results}

\begin{figure*}[t!]
    \centering
    \includegraphics[width=0.93\textwidth]{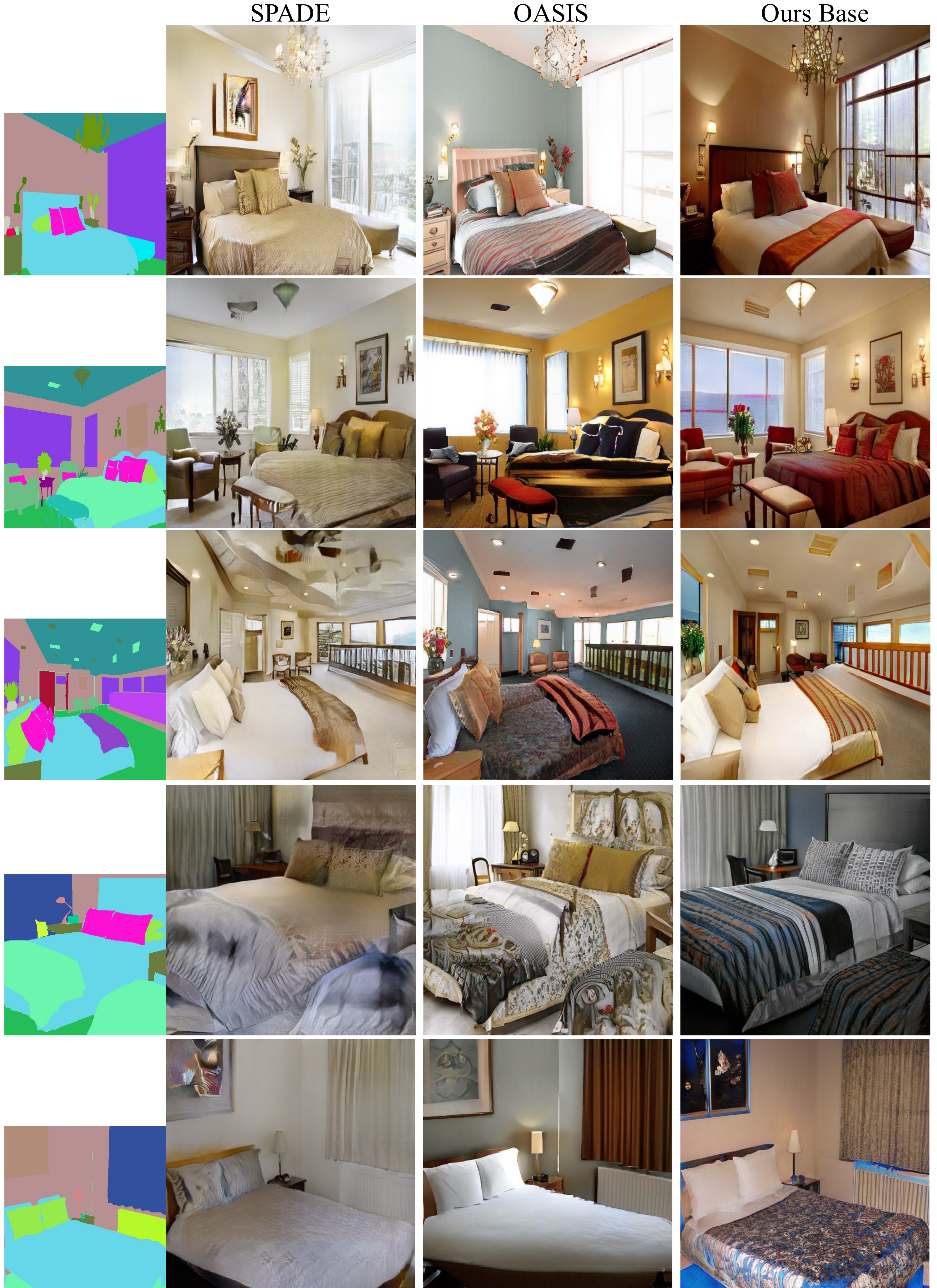}
    \caption{Visual comparisons of synthesis results by different methods ($512\times512$) on the ADE20K bedroom.}
    \label{fig:baselne_bedroom}
    \vspace{-0.1in}
\end{figure*}

\begin{figure*}[t!]
    \centering
    \includegraphics[width=0.93\textwidth]{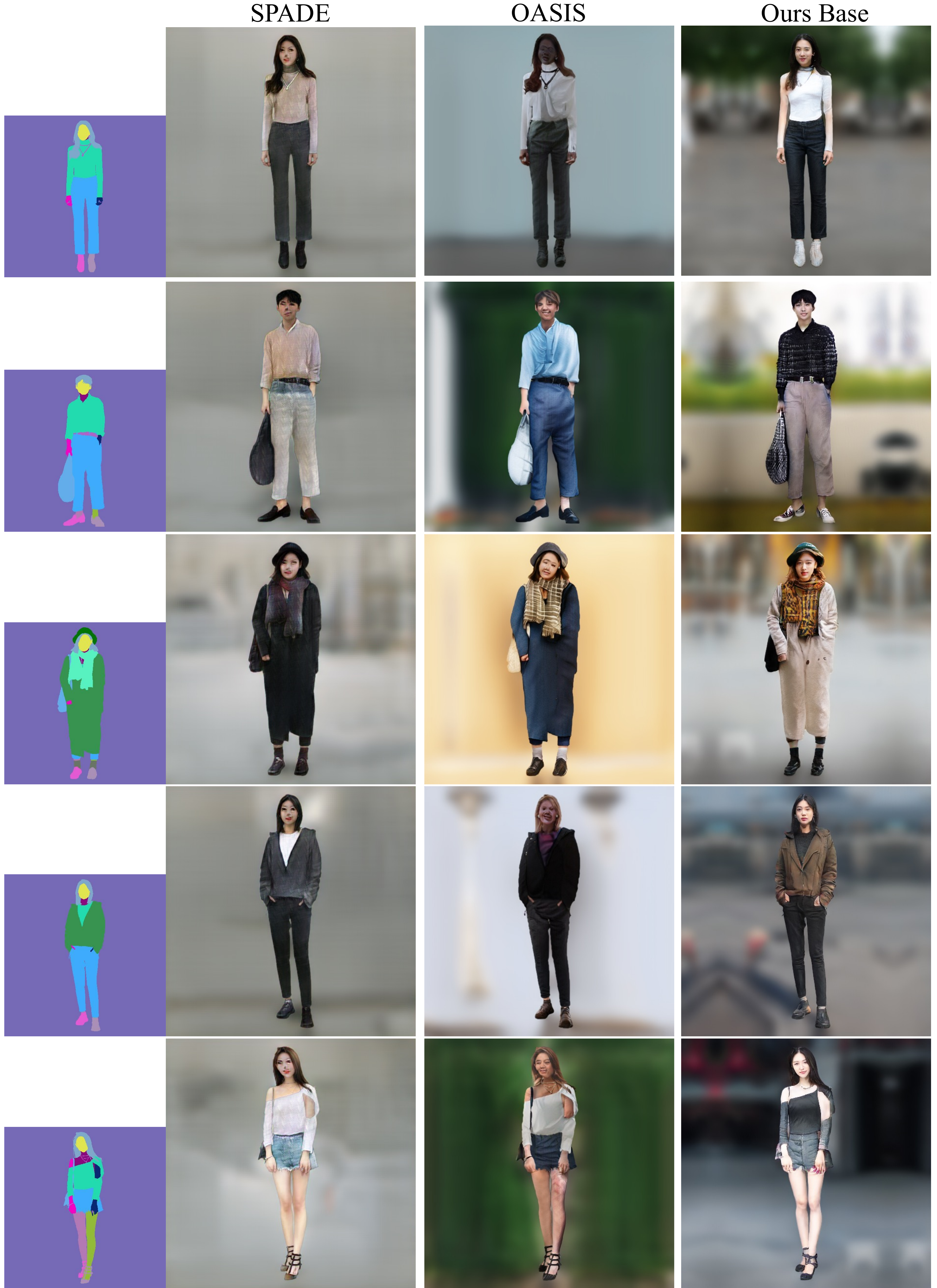}
    \caption{Visual comparisons of synthesis results by different methods ($512\times512$) on the full human body dataset.}
    \label{fig:baseline_human}
    \vspace{-0.1in}
\end{figure*}

\begin{figure*}[t!]
    \centering
    \includegraphics[width=0.93\textwidth]{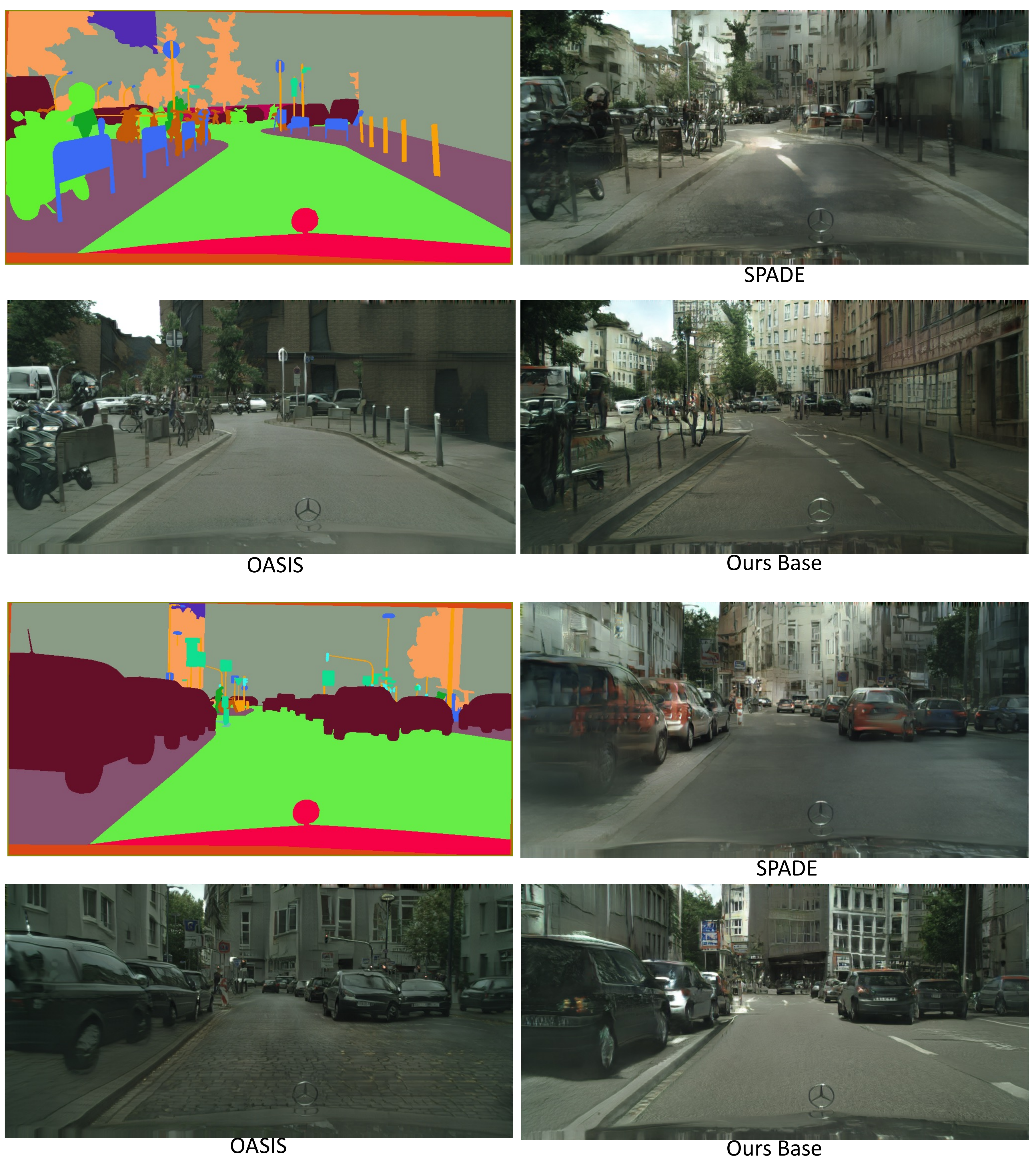}
    \caption{Visual comparisons of synthesis results by different methods ($1024\times512$) on the Cityscapes dataset.}
    \label{fig:baseline_city}
    \vspace{-0.1in}
\end{figure*}

\begin{figure*}[t!]
    \centering
    \includegraphics[width=0.93\textwidth]{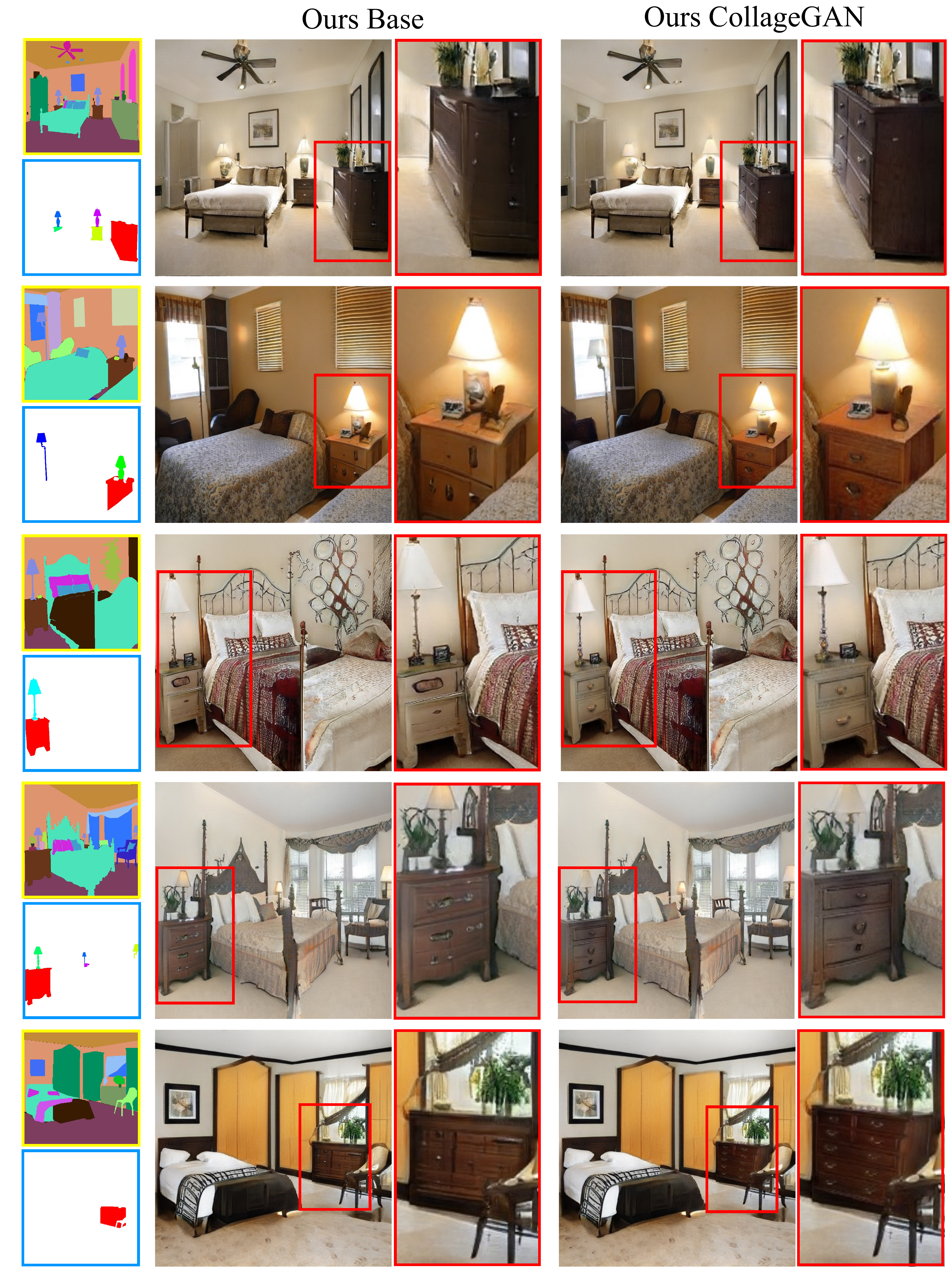}
    \caption{Comparisons between our base model and our CollageGAN model on the ADE20k bedroom.}
    \label{fig:composition_bedroom}
    \vspace{-0.1in}
\end{figure*}

\begin{figure*}[t!]
    \centering
    \includegraphics[width=0.93\textwidth]{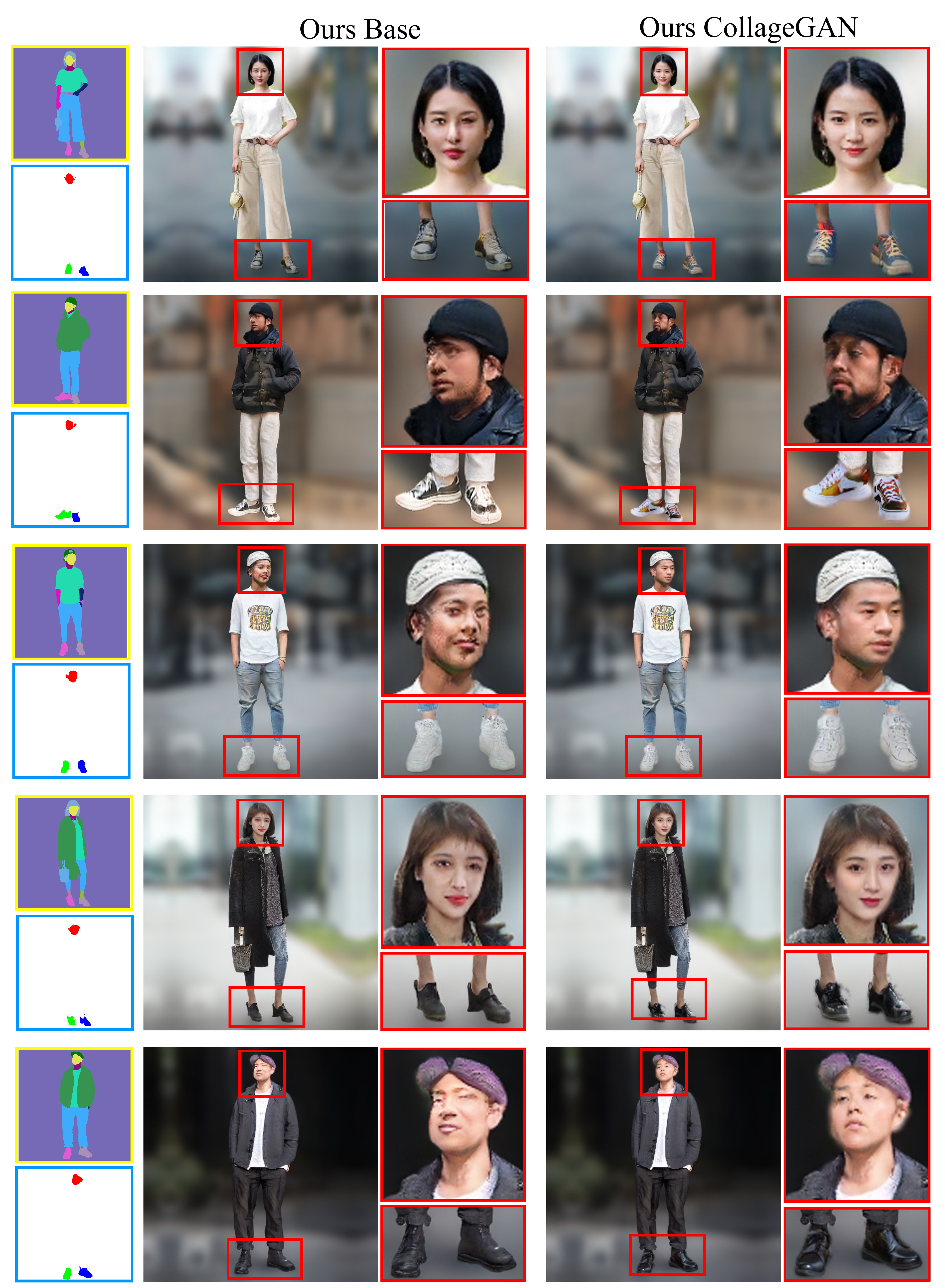}
    \caption{Comparisons between our base model and our CollageGAN model on the full human body dataset.}
    \label{fig:composition_human}
    \vspace{-0.1in}
\end{figure*}

\begin{figure*}[t!]
    \centering
    \includegraphics[width=0.93\textwidth]{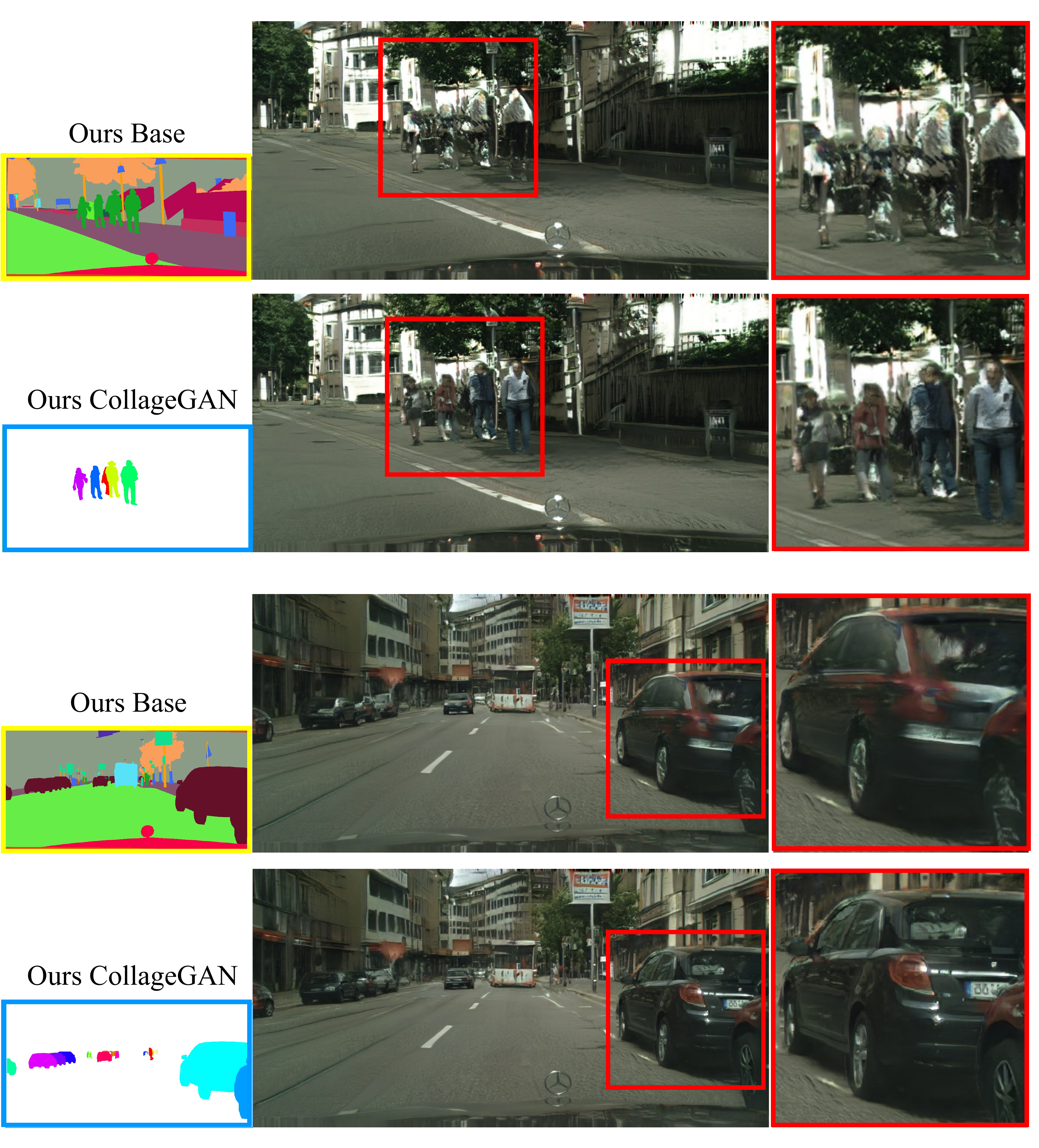}
    \caption{Comparisons between our base model and our CollageGAN model on the Cityscapes dataset.}
    \label{fig:composition_city}
    \vspace{-0.1in}
\end{figure*}

\begin{figure*}[t!]
    \centering
    \includegraphics[width=0.95\textwidth]{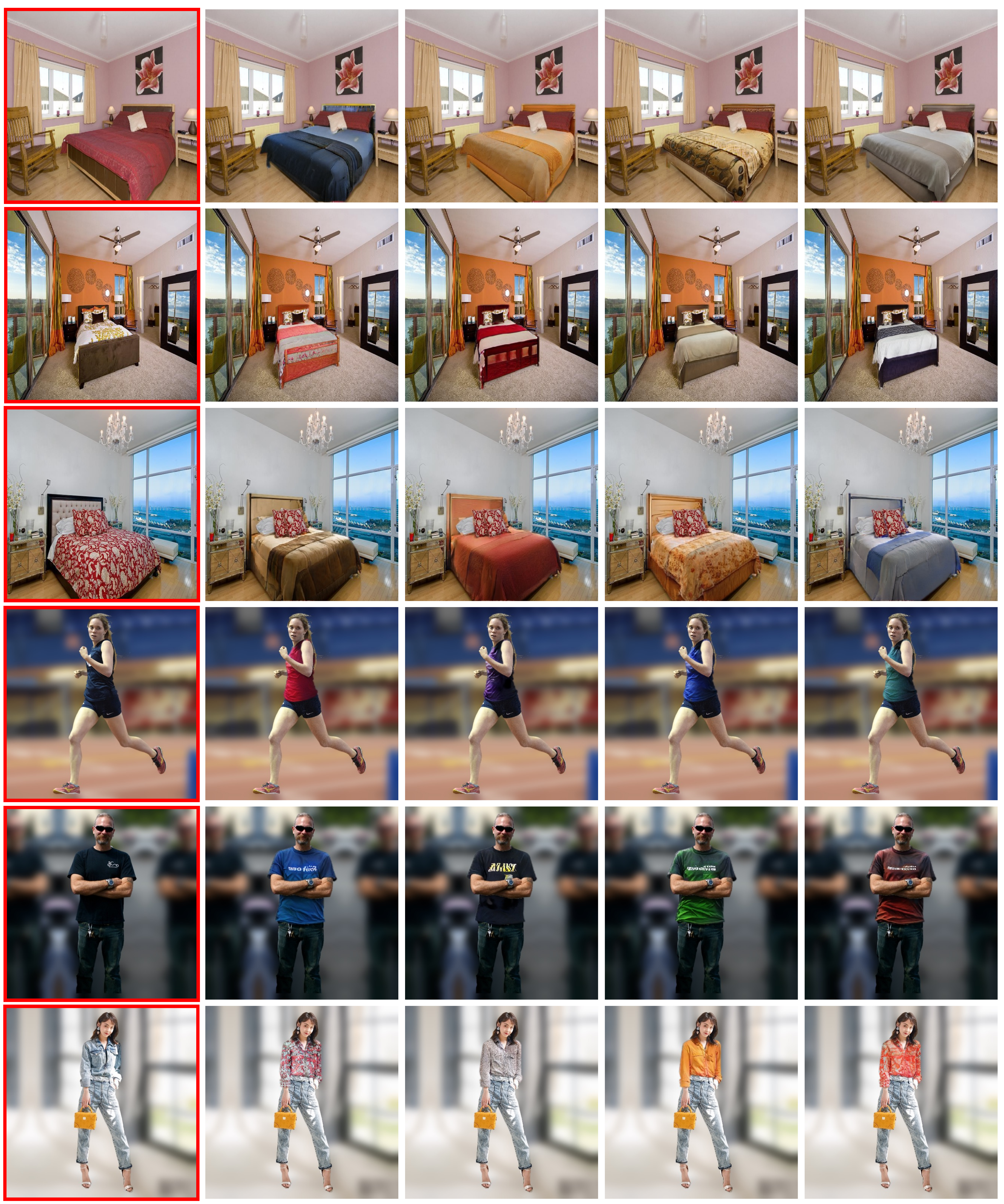}
    \caption{Images in the red box are real images. Here we use the \textbf{bed} (top three rows) and \textbf{uppercloth} (bottom three rows) specific generator to replace the original objects.}
    \label{fig:supp_replace_real}
    \vspace{-0.1in}
\end{figure*}

\begin{figure*}[t!]
    \centering
    \includegraphics[width=0.95\textwidth]{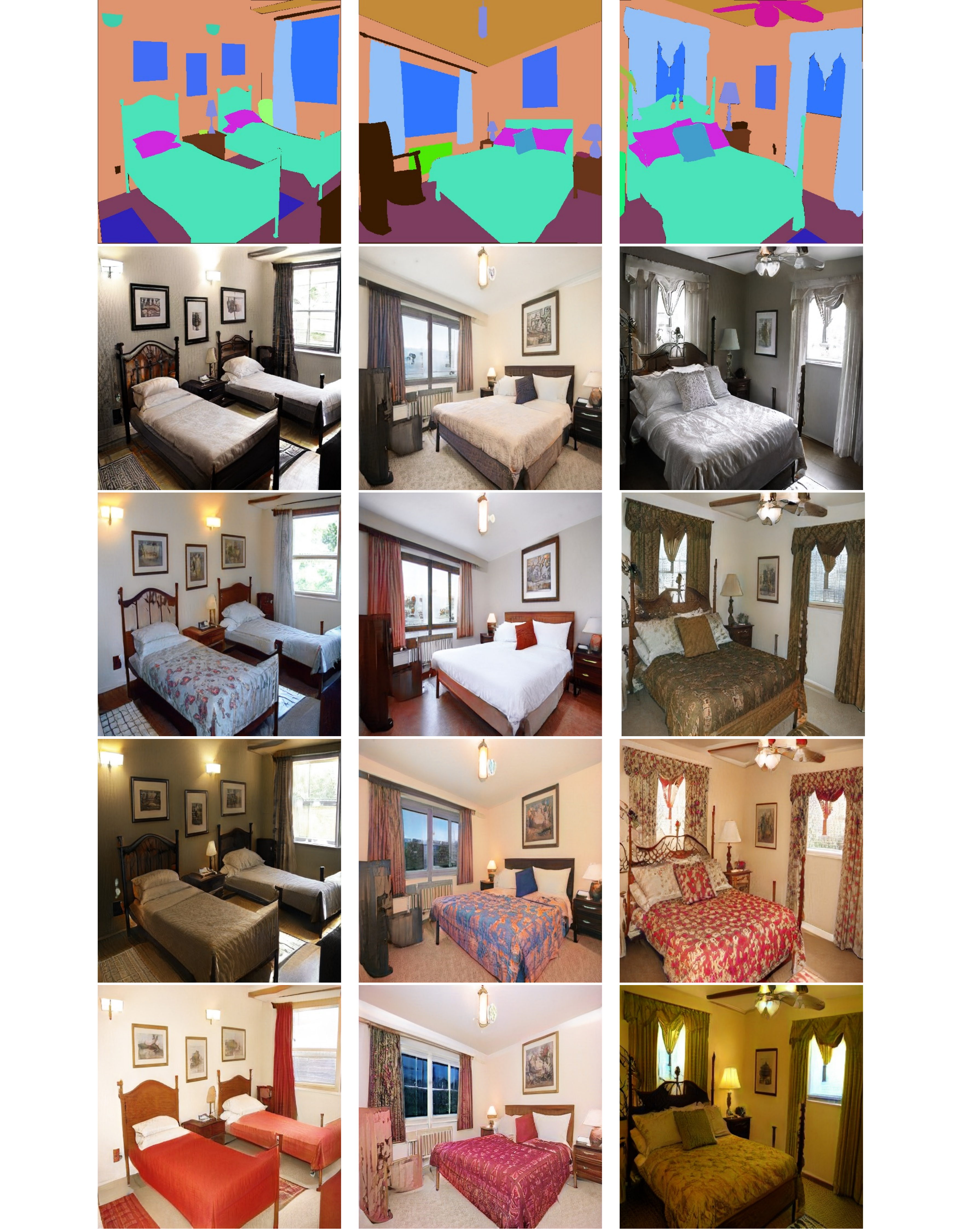}
    \caption{Multi-modal synthesis results on ADE20 bedroom by our base model. Each column shows multiple generations for same semantic mask.}
    \label{fig:multi_bedroom}
    \vspace{-0.1in}
\end{figure*}

\begin{figure*}[t!]
    \centering
    \includegraphics[width=0.95\textwidth]{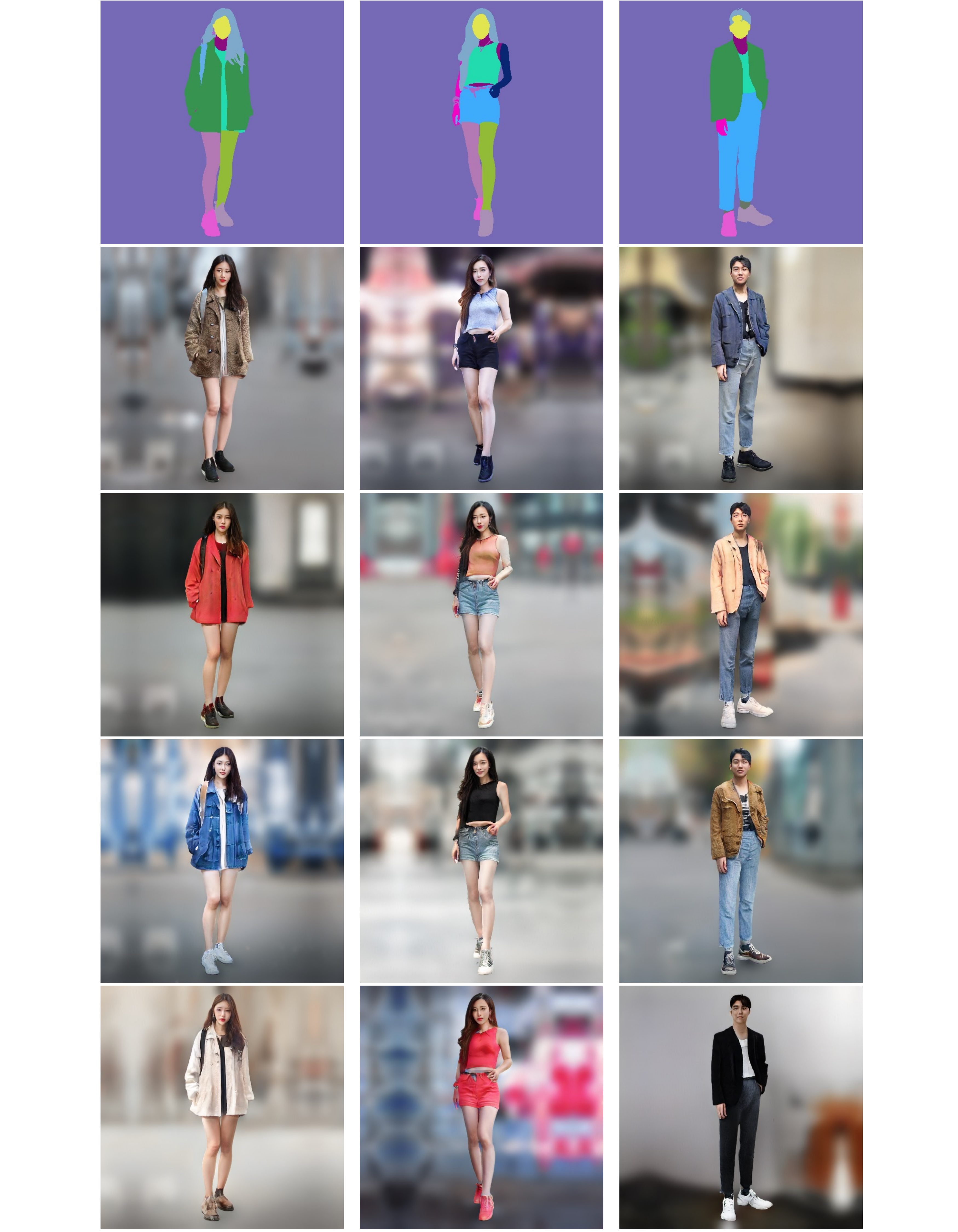}
    \caption{Multi-modal synthesis results on full human body by our base model. Each column shows multiple generations for same semantic mask.}
    \label{fig:multi_human}
    \vspace{-0.1in}
\end{figure*}

\begin{figure*}[t!]
    \centering
    \includegraphics[width=0.95\textwidth]{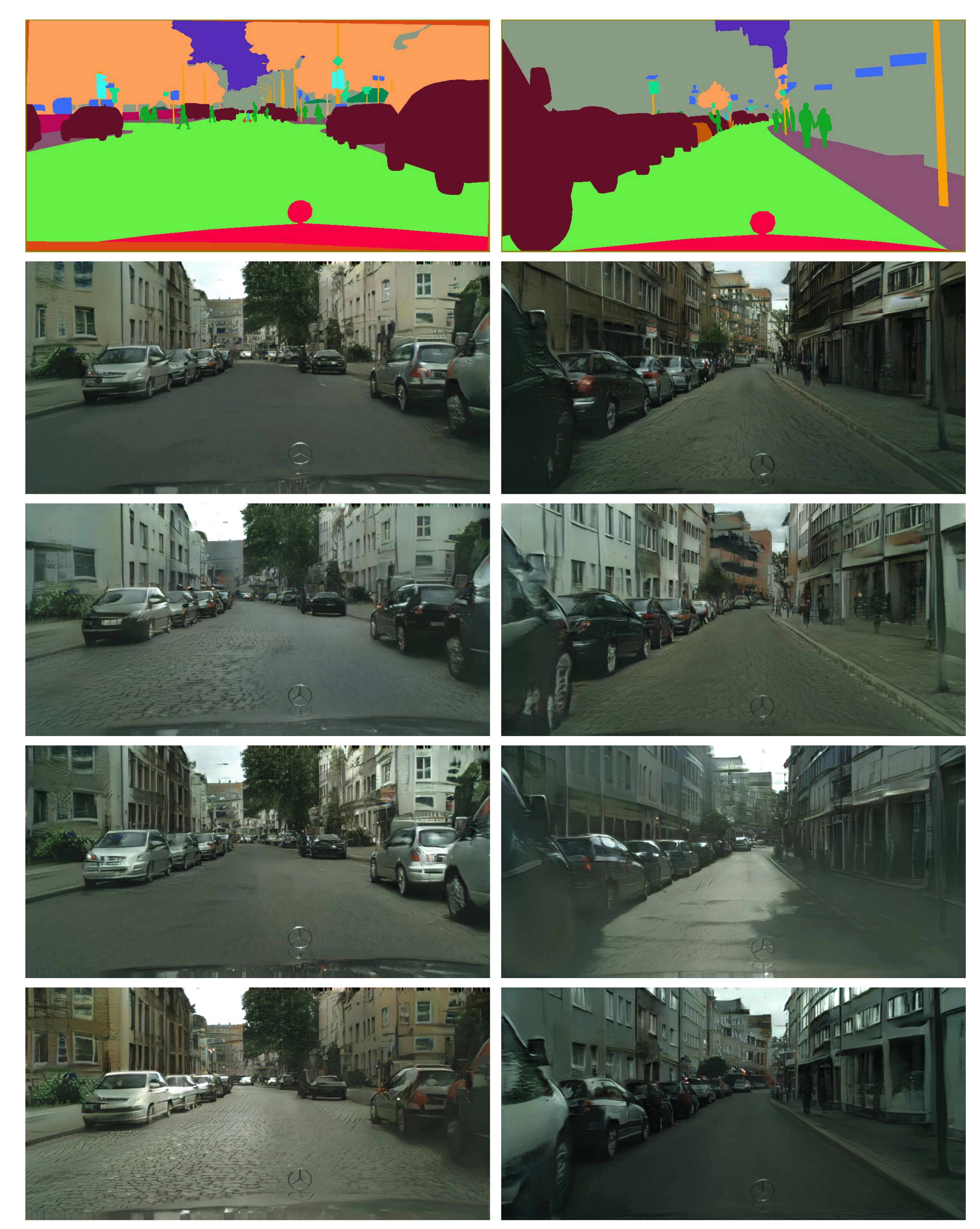}
    \caption{Multi-modal synthesis results on cityscapes by our base model. Each column shows multiple generations for same semantic mask.}
    \label{fig:multi_city}
    \vspace{-0.1in}
\end{figure*}

\begin{figure*}[t!]
    \centering
    \includegraphics[width=0.95\textwidth]{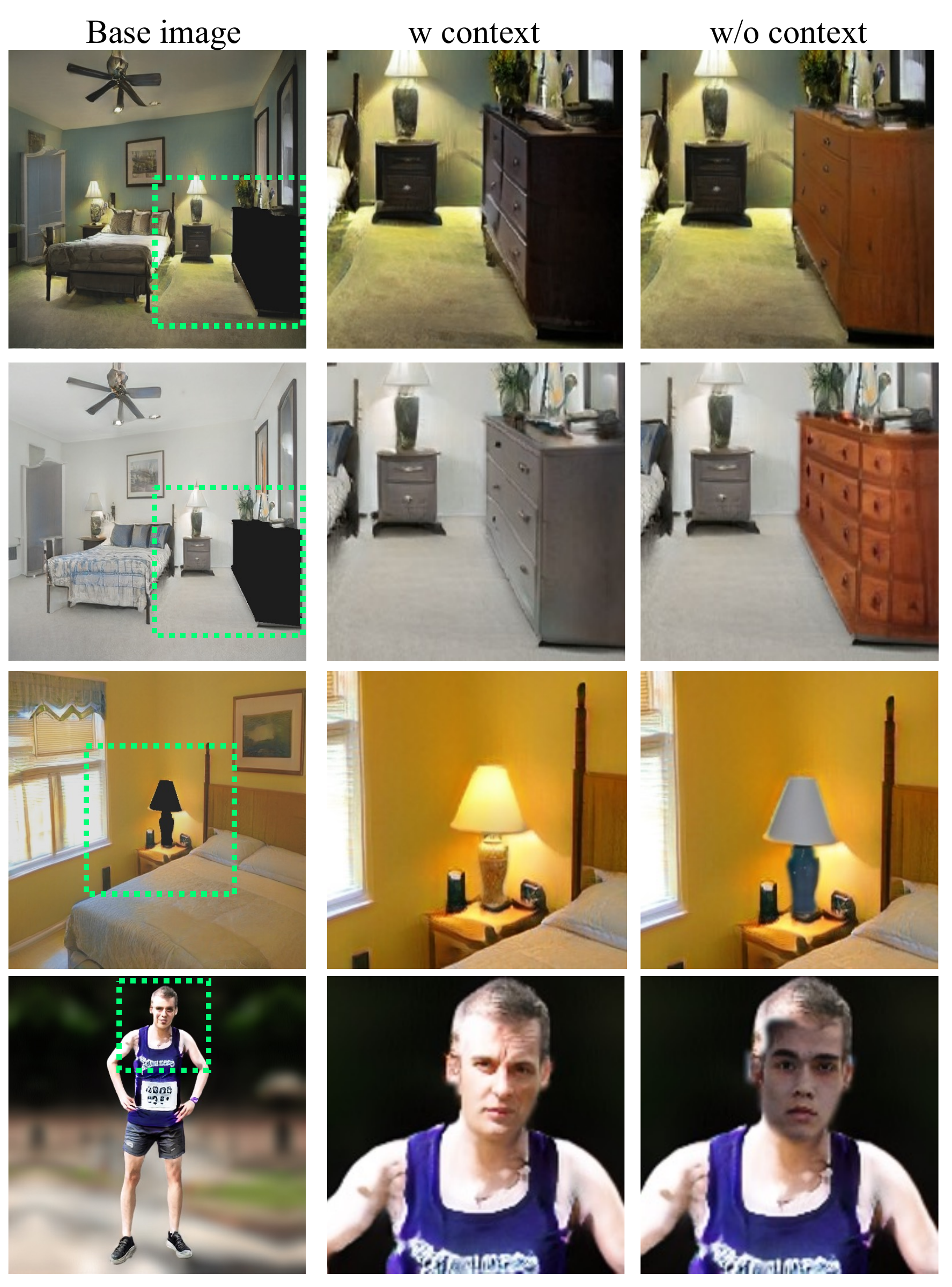}
    \caption{class-specific generator will generate inconsistent results without context.}
    \label{fig:context}
    \vspace{-0.1in}
\end{figure*}

Fig.~\ref{fig:baselne_bedroom}$-$\ref{fig:baseline_city} show the our base model comparison with baselines on different datasets.  Fig.~\ref{fig:composition_bedroom}$-$\ref{fig:composition_city} show the comparison between our base model results and composition results using class-specific models. Fig.~\ref{fig:supp_replace_real} shows more results of replacing real objects in the real image. Please \textbf{zoom in} and check how well our class-specific generator can generate instances that look consistent with their background, such as the lighting condition on the beds and upperclothes in the 4th and 5th rows.  Fig.~\ref{fig:multi_bedroom}$-$\ref{fig:multi_city} shows our base model multi-modal ability on semantic to image generation. Here we sample different $z$ code to get different results. Fig.~\ref{fig:context} shows importance of context for our class-specific generator. For example, the models may lack knowledge of orientation (row 1,2) or lighting condition (row 1,3,4) without context.   

\begin{figure*}[t!]
    \centering
    \includegraphics[width=0.95\textwidth]{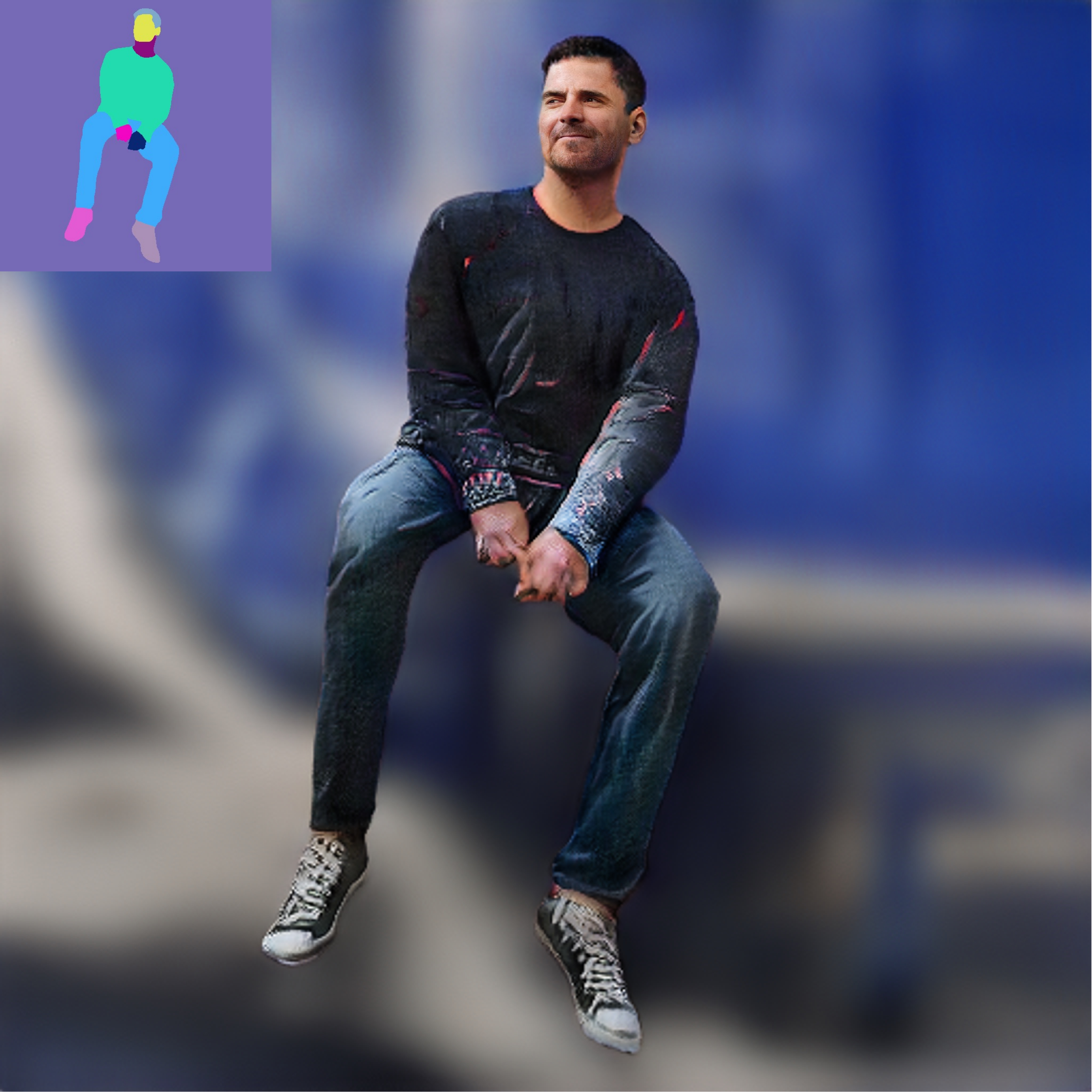}
    \caption{The mixed-resolution result. Here the base image is first resized to $4096\times4096$ and then face region is composited by high quality face generated from face specific model.}
    \label{fig:mixed1}
    \vspace{-0.1in}
\end{figure*}

\begin{figure*}[t!]
    \centering
    \includegraphics[width=0.95\textwidth]{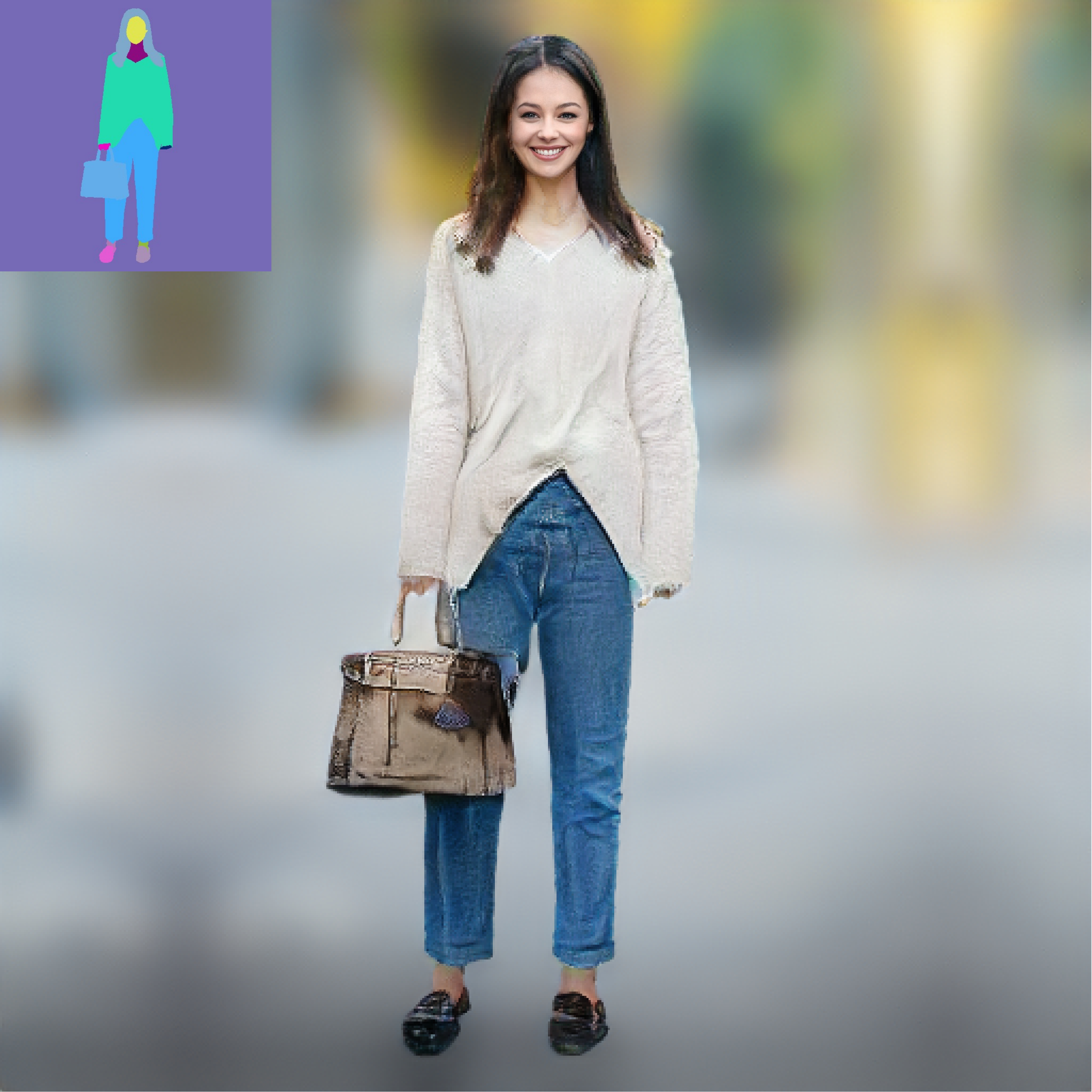}
    \caption{The mixed-resolution result. Here the base image is first resized to $4096\times4096$ and then face region is composited by high quality face generated from face specific model.}
    \label{fig:mixed2}
    \vspace{-0.1in}
\end{figure*}

\begin{figure*}[t!]
    \centering
    \includegraphics[width=0.93\textwidth]{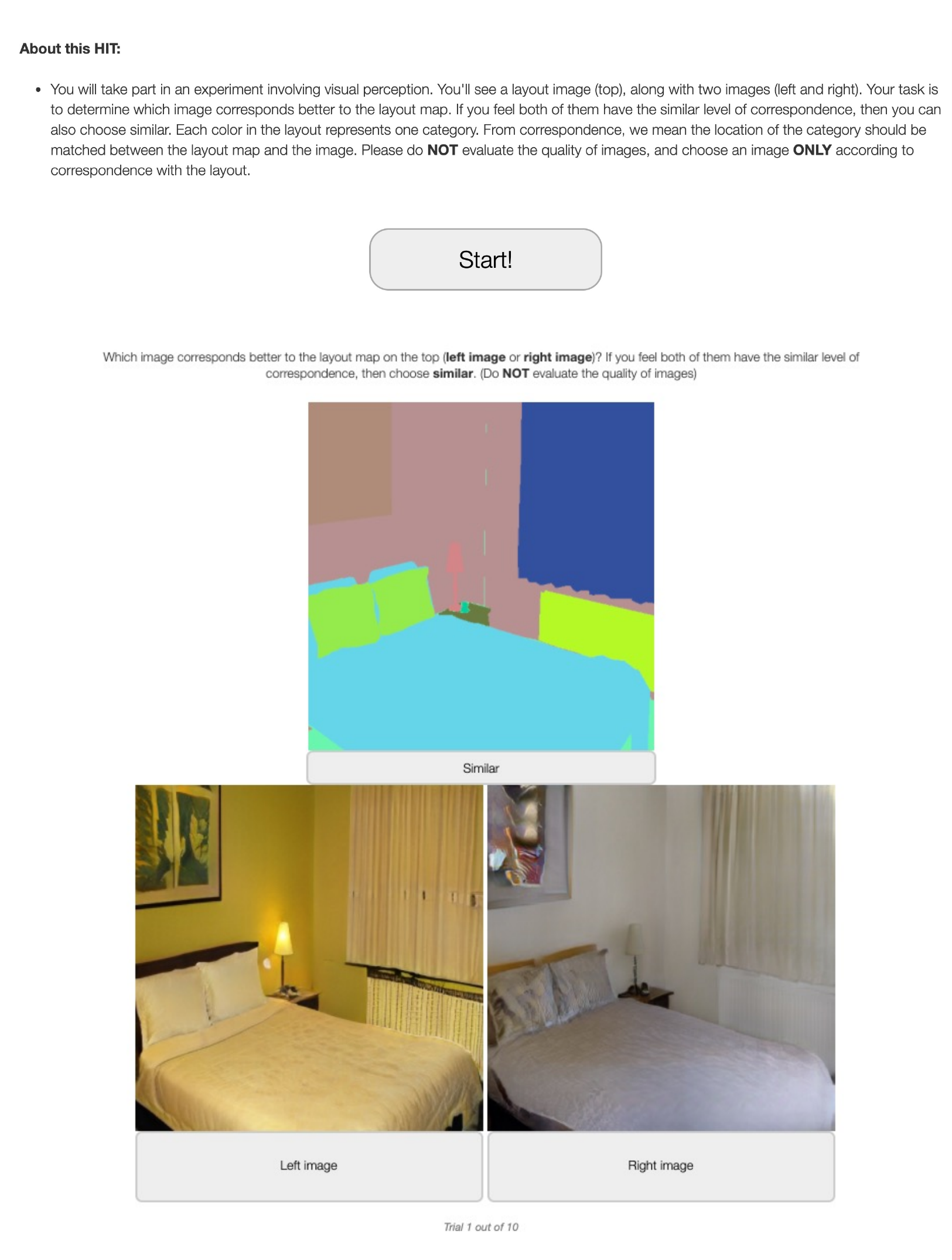}
    \caption{Alignment user study screenshot.}
    \label{fig:user_correspondence}
    \vspace{-0.1in}
\end{figure*}

\begin{figure*}[t!]
    \centering
    \includegraphics[width=0.93\textwidth]{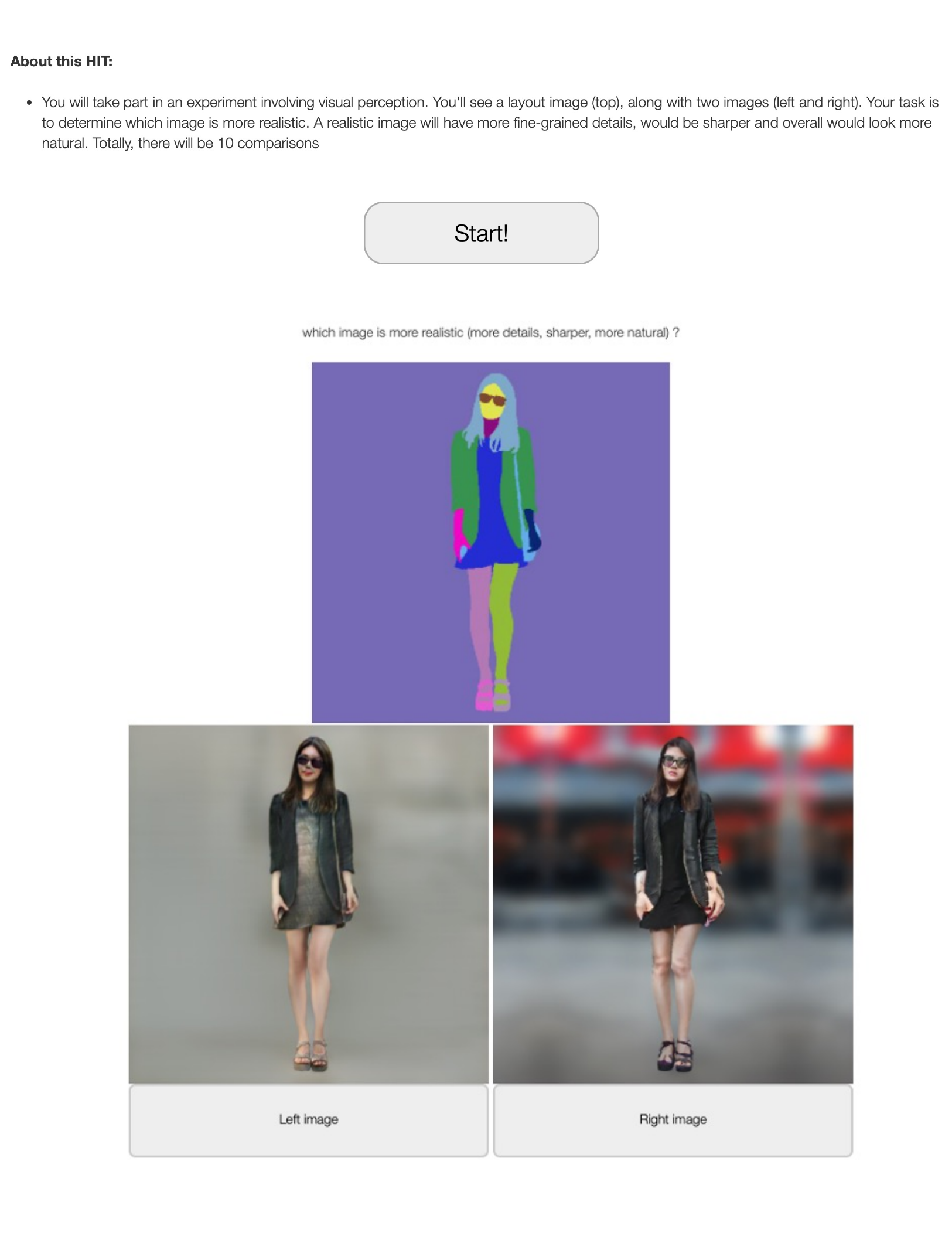}
    \caption{Realism user study screenshot.}
    \label{fig:user_realism}
    \vspace{-0.1in}
\end{figure*}

\begin{figure*}[t!]
    \centering
    \includegraphics[width=0.93\textwidth]{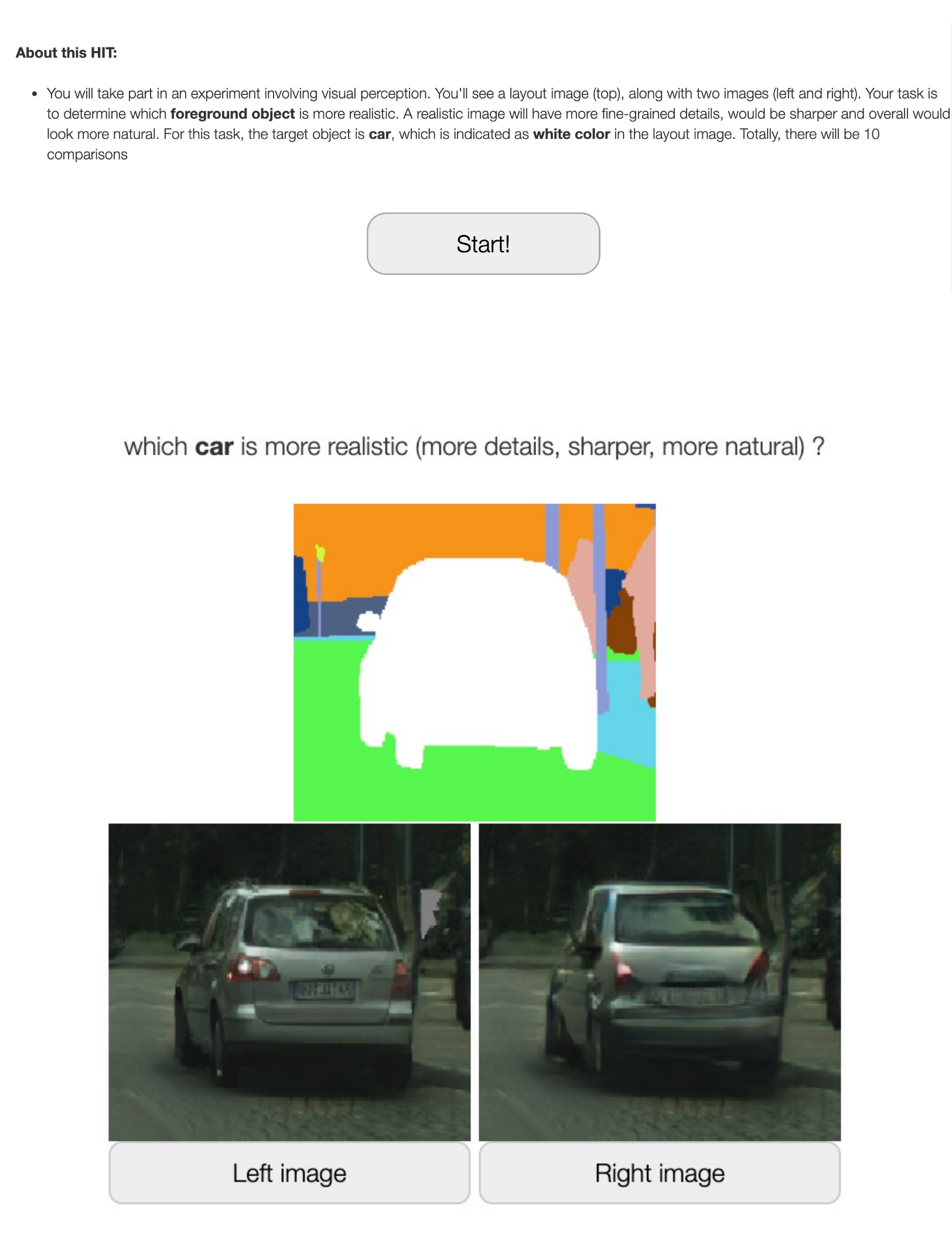}
    \caption{Instance realism user study screenshot.}
    \label{fig:user_instance}
    \vspace{-0.1in}
\end{figure*}

Finally, we show another interesting application of our class-specific model in Fig.~\ref{fig:mixed1}$-$\ref{fig:mixed2} to generate mixed-resolution results where the important region is in high resolution. Specifically, we train a face generator (with blurred inputs) using the CelebA dataset. Different from the previously mentioned face generator, this model is trained on the $256\times256$ resolution. When we do the composition, we first use our base model to generate a full human body base image $I_b$ and resize this image to $4096\times4096$ resolution. Then the face region can be replaced by the high quality face $I_c$ generated from the class-specific face model.  One interesting observation is that though the majority ethnicity in our collected full human body data is Asian, we find more western faces in the mixed resolution results due to the different data distribution in the CelebA dataset. Thus, our class-specific model idea can also be used to reduce bias and increase the diversity of certain classes during image synthesis by levaraging data from different sources.

\end{document}